\documentclass{article}

\usepackage{arxiv}

\usepackage[utf8]{inputenc} 
\usepackage[T1]{fontenc}    
\usepackage{hyperref}       
\usepackage{url}            
\usepackage{booktabs}       
\usepackage{amsfonts}       
\usepackage{nicefrac}       
\usepackage{microtype}      
\usepackage{lipsum}
\usepackage{graphicx}
\usepackage[algo2e]{algorithm2e}
\usepackage{algorithm}
\usepackage{algpseudocode}
\usepackage[countmax]{subfloat}
\usepackage{amsmath}
\usepackage{subcaption}
\usepackage{amssymb} 
\usepackage{pifont}  
\usepackage{multirow}
\usepackage{bbding}
\graphicspath{ {./images/} }

\usepackage{multirow}
\usepackage{bbding}
\usepackage[algo2e]{algorithm2e}
\usepackage{algorithm}
\usepackage{algpseudocode}
\usepackage[countmax]{subfloat}
\usepackage{amsthm}
\newtheorem{theorem}{Theorem}
\usepackage{wrapfig}
\usepackage{xcolor}
\usepackage{makecell}
\usepackage{bm}
\usepackage{amsfonts}
\usepackage{amsmath}
\usepackage{caption}
\usepackage{float}
\usepackage{paracol} 
\usepackage{booktabs}
\usepackage{lipsum}

\title{Hyperbolic Cycle Alignment for Infrared-Visible Image Fusion}

\author{
 Timing Li \\
  College of Intelligence and Computing\\
  Tianjin University\\
  \texttt{litm@tju.edu.cn} \\
   \And
 Bing Cao \\
  College of Intelligence and Computing\\
  Tianjin University\\
  \texttt{caobing@tju.edu.cn} \\
  \And
 Jiahe Feng \\
  College of Intelligence and Computing\\
  Tianjin University\\
  \And
 Haifang Cao \\
  College of Intelligence and Computing\\
  Tianjin University\\
  \And
 Qinghua Hu \\
  College of Intelligence and Computing\\
  Tianjin University\\
  \texttt{huqinghua@tju.edu.cn} \\
  \And
  Pengfei Zhu \\
  College of Intelligence and Computing\\
  Tianjin University\\
  \texttt{zhupengfei@tju.edu.cn} \\
}

\begin{document}
\maketitle
\begin{abstract}
Image fusion synthesizes complementary information from multiple sources, mitigating the inherent limitations of unimodal imaging systems. Accurate image registration is essential for effective multi-source data fusion. However, existing registration methods, often based on image translation in Euclidean space, fail to handle cross-modal misalignment effectively, resulting in suboptimal alignment and fusion quality. To overcome this limitation, we explore image alignment in non-Euclidean space and propose a Hyperbolic Cycle Alignment Network (Hy-CycleAlign). To the best of our knowledge, Hy-CycleAlign is the first image registration method based on hyperbolic space. It introduces a dual-path cross-modal cyclic registration framework, in which a forward registration network aligns cross-modal inputs, while a backward registration network reconstructs the original image, forming a closed-loop registration structure with geometric consistency. Additionally, we design a Hyperbolic Hierarchy Contrastive Alignment (H$^{2}$CA) module, which maps images into hyperbolic space and imposes registration constraints, effectively reducing interference caused by modality discrepancies. We further analyze image registration in both Euclidean and hyperbolic spaces, demonstrating that hyperbolic space enables more sensitive and effective multi-modal image registration. Extensive experiments on misaligned multi-modal images demonstrate that our method significantly outperforms existing approaches in both image alignment and fusion. Our code will be publicly available.
\end{abstract}


\section{Introduction}
\label{sec:intro}

Multi-modal image fusion integrates complementary information from heterogeneous sensors and has become a key technology for enhancing visual perception and analytical capabilities across various domains. By integrating the advantages of different modalities, such as the thermal radiation sensitivity of infrared imaging and the high-resolution texture details of visible, fusion technology generates comprehensive scene representations, enabling applications ranging from all-weather surveillance to search and rescue in harsh environments. 

The key to the success of such a fusion system lies in accurate multi-modal image alignment, a process that establishes pixel-level spatial correspondences between modalities. Even minor misalignments, such as a displacement of thermal signatures in infrared images relative to visible edges, can lead to ghosting artifacts or misinterpretations. For example, in security surveillance, an unregistered fusion of infrared and visible feeds might erroneously overlay a human heat signature onto a nearby object, compromising threat identification accuracy. Despite its importance, achieving reliable registration between infrared and visible images remains a challenge.

The primary causes of multi-modal image misalignment include several factors. The positions of different sensors cannot be perfectly identical, resulting in displacement deviations in the captured images. Secondly, infrared imaging relies on thermal radiation while visible imaging relies on light reflection, and this difference in imaging mechanisms can lead to the mismatch of edge features. Additionally, complex factors such as viewpoint changes, motion blur, and rotational variations in real-world dynamic scenes further exacerbate the misalignment problem in multi-modal images. Unfortunately, mainstream fusion algorithms typically assume that the input images are pre-aligned, overlooking the inherent connection between registration and fusion. This idealized assumption limits their adaptability to partially aligned or noisy multi-modal data. Considering the difference in imaging mechanisms, the misalignment between infrared and visible images represents a nonlinear disparity, which is further exacerbated in dynamic scenes. 

To address the difficulty of cross-modal pixel alignment of infrared and visible images in Euclidean space, we have realized pixel-level multimodal image alignment in hyperbolic space for the first time. We propose a Hyperbolic Space-based Cyclic Consistency Alignment Network to realize it, termed as Hy-CycleAlign. The main contributions are summarized as follows:

\begin{itemize}
\item A Hyperbolic Space-based Cyclic Consistency Alignment Network is proposed, which introduces a dual-path cross-modal cyclic registration framework. By coordinating registration and inverse registration, the framework establishes a closed-loop registration structure.
\item We introduce a Hyperbolic Hierarchical Contrastive Alignment, which first maps the input images into Poincar{\'e} space to guide the alignment process within hyperbolic geometry.  This design effectively mitigates the impact of nonlinear cross-modal discrepancies by leveraging the structural properties of hyperbolic space.
\item We provide a theoretical analysis demonstrating that hyperbolic space, particularly within the Poincar{\'e} model, exhibits greater sensitivity to distance variations.  Extensive experiments on misaligned infrared-visible image fusion tasks validate the effectiveness of our method.
\end{itemize}

\begin{figure}[t]
\centering
\includegraphics[width=1.\columnwidth]{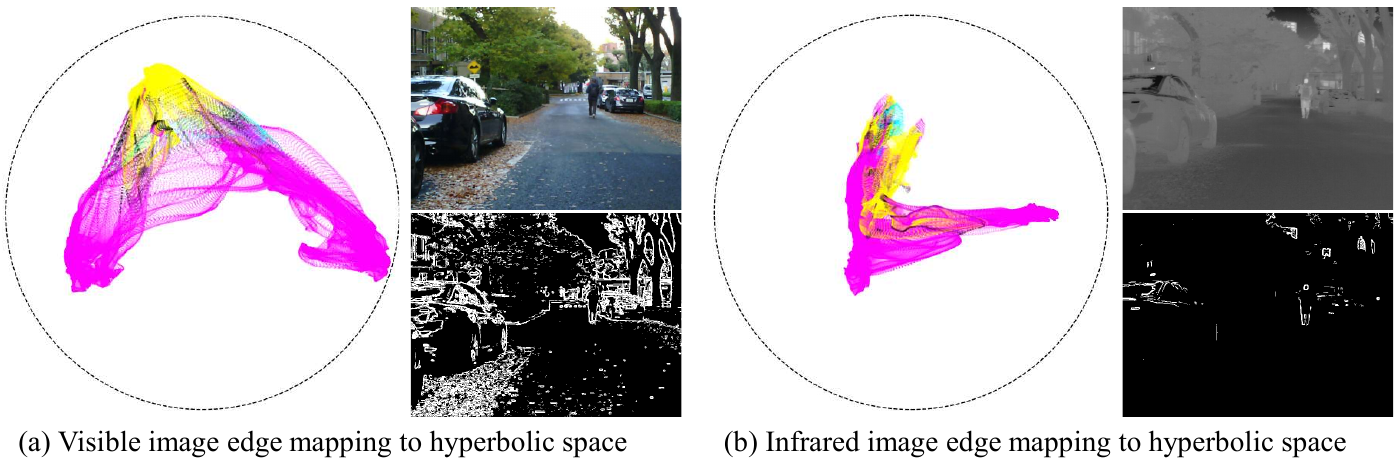} 
\caption{Comparison of different modalities in Euclidean and hyperbolic spaces: the background is shown in magenta, humans in cyan, vehicles in yellow, and object edges in black. In Euclidean space, edge maps tend to appear more scattered and lack hierarchical structure, whereas in hyperbolic space, edge features exhibit more pronounced clustering and hierarchy.}
\label{first}
\end{figure}

\section{Related Works}
This section briefly reviews representative deep learning-based methods for multi-modal image alignment and fusion, as well as relevant foundational studies on non-Euclidean space.

\subsection{Infrared-visible images alignment and fusion}
These fusion methods fundamentally depend on the availability of strictly aligned input images. Any misalignment between the source images can significantly degrade the quality of the fused output. Broadly, these techniques can be categorized into three main types based on their underlying architectures: convolutional neural network (CNN)-based methods\cite{zhang2020ifcnn}, hybrid CNN-Transformer-based methods\cite{xu2020u2fusion}, and Generative Adversarial Network (GAN)-based methods\cite{zhao2020didfuse}. These methods improve the final fusion results through a series of carefully designed components, including feature extraction modules, reconstruction modules, and fusion modules.

Fusion methods based on unaligned images are specifically designed to address misalignment and image degradation issues that arise when existing approaches attempt to fuse unaligned image pairs. In recent years, to better achieve fusion of misaligned images, several studies such as ReCoNet \cite{huang2022reconet}, SuperFusion \cite{TANG2022SuperFusion_}, MURF \cite{xu2023murf}, UMF \cite{chen2022unsupervised}, and IMF \cite{wang2024improving} have been conducted to address this challenge. In existing registration-fusion methods rely on image translation modules which not only introduce additional noise but also make image registration heavily dependent on the performance of the translation modules. Therefore, how to improve image registration performance without introducing additional noise remains a critical challenge to be addressed.

\subsection{Hyperbolic Deep Learning}
Due to its negative curvature, hyperbolic space can more efficiently capture hierarchical and tree-like structures in data. Therefore, it has been widely adopted in fields such as graphs \cite{bachmann2020constant, chami2019hyperbolic, dai2021hyperbolic, cao2025hyperbolic, liu2019hyperbolic, lou2020differentiating}, text \cite{aly2020every, tifrea2018poincar, zhu2020hypertext, ramasinghe2024accept, yang2024hypformer}, and vision \cite{mettes2024hyperbolic, li2024hyperbolic, khrulkov2020hyperbolic, atigh2022hyperbolic} tasks to address the limitations of Euclidean space in modeling hierarchical data. In this work, we build upon these foundations and take a step toward pixel-level multi-modal image registration by applying constraints in hyperbolic space to reduce nonlinear modality discrepancies.

Existing research has already demonstrated the substantial potential of hyperbolic space in effectively handling problems characterized by hierarchical structures. GhadimiAtigh et al. \cite{atigh2022hyperbolic} verified that embedding pixels into hyperbolic space can accurately map the interiors and edges of an object, thereby achieving precise image segmentation. Khrulkov et al. \cite{khrulkov2020hyperbolic} treated image degradation as a hierarchy, embedding it into hyperbolic space to achieve better re-identification performance. Fu et al. \cite{fu2024cf} introduced hyperbolic space to the object detection task and achieved weak alignment at the feature level. Li et al. \cite{li2024hyperbolic} used hyperbolic distance metrics to represent the distance between features, enabling anomaly detection in hyperbolic space. 

Although hyperbolic space has shown excellent performance in various vision tasks, most of these tasks focus on feature-level processing and unimodal vision tasks. To date, there has been no research on pixel-level image registration based on hyperbolic space. To fill this gap, we explore the image registration problem in hyperbolic space and propose a hyperbolic space-based pixel-level alignment method. This approach breaks through the nonlinearity limitations of traditional registration methods in Euclidean space and achieves promising results.

\section{Method}
\label{sec: method}
\begin{figure}[t]
\centering
\includegraphics[width=1.\columnwidth]{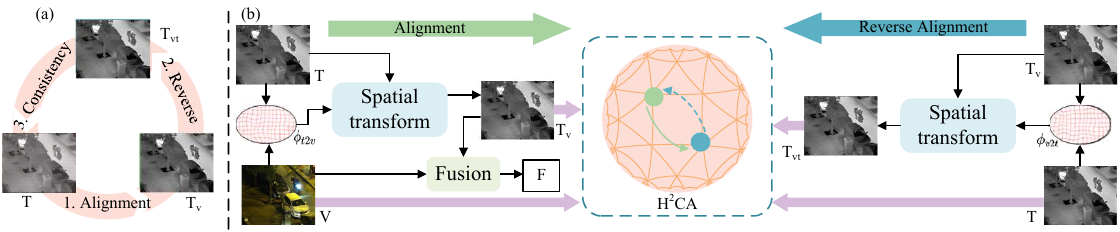} 
\caption{Overview of the proposed method. (a) is the Hy-CycleAlign alignment process in which the Hyperbolic Hierarchy Contrastive Alignment (H$^{2}$CA) module aligns the infrared image to the visible image, followed by re-aligning the result back to the original image. The H$^{2}$CA maps the images into the Poincar{\'e} space and constraints, thereby achieving effective infrared-visible image registration.}
\label{arc}
\end{figure}

This section presents our Hy-CycleAlign method, a cyclic consistency alignment model based on hyperbolic space, as shown in Fig. \ref{arc}. We first analyze the advantages of constraining multi-modal image registration in hyperbolic space. Then, we introduce the cycle-consistent registration framework. Finally, we propose a hierarchical registration constraint in hyperbolic space, which enables pixel-level alignment and fusion of infrared and visible images.

\subsection{Motivation}
\label{sec: motivation}
Due to the modality differences between infrared and visible images \cite{brown1992survey}, the mapping relationship between them is nonlinear, making it difficult to impose accurate alignment constraints within Euclidean space. Although image translation networks are widely used in modern image registration tasks, they often lead to structural distortions \cite{kong2021breaking, chen2022unsupervised}, loss of semantic information \cite{wu2021discover}, and accumulation of indirect errors \cite{fan2024gls, xiang2022infrared}. Considering that multi-modal images registration is a nonlinear problem, and inspired by works such as \cite{tifrea2018poincar, li2024hyperbolic, atigh2022hyperbolic} that leverage hyperbolic space to handle nonlinearities, we explore the impact of hyperbolic space on multi-modal image registration.

Images contain implicit hierarchical structural information, which can be better represented in hyperbolic space to capture their hierarchical structure and complex relationships \cite{itti2002model, li2023efficient, atigh2022hyperbolic}. Considering that the Poincar{\'e} space has the characteristic of negative curvature, which makes its spatial tree-like expansion structure more naturally realize the mapping of 2-dimensional graphs to the hyperbolic space, we choose the Poincar{\'e} space to carry out the study of multi-modal image alignment on the hyperbolic space \cite{pal2024compositional}.

\begin{theorem}
Compared to Euclidean space, the hyperbolic space represented by the Poincar{\'e} space is more sensitive to misalignments, and this sensitivity increases as points approach the boundary of the Poincar{\'e} space.
\end{theorem}

\begin{proof}
Assuming $u$ and $v$ are the points to be registered from different modality images, their distance in Euclidean space $d_{E}(u,v)$ can be expressed as
\begin{equation}
\label{De_dist}
d_{E}(u,v)=\left \| u-v \right \| _{2}.
\end{equation}

Assuming the normal Poincar{\'e} space $\mathbb{D}^{n}=\{ x\in \mathbb{R}^{n}:\left \| x \right \| < 1 \}$, $x$ denotes a point in the Poincar{\'e} space. Then, the distance $d_{p}(u,v)$ between points u and v in the Poincar{\'e} space is shown as
\begin{equation}
\label{Dp_dist}
d_{P}(u,v)=\cosh^{-1} (1+2\frac{\left \| u-v \right \|^{2}}{(1-\left \| u \right \|^{2})(1-\left \| v \right \|^{2})} ).
\end{equation}

Let $\delta =v-u$, in the alignment task, the goal is to make $v\to u$. Then, it follows that $\left \| v \right \| \approx  \left \| u \right \| $.

We define X in Eq. \ref{Def_X}, 
\begin{equation}
\label{Def_X}
X=2\frac{\left \| u-v \right \|^{2}}{(1-\left \| u \right \|^{2})(1-\left \| v \right \|^{2})}\approx 2\frac{\left \| \delta  \right \|^{2}}{(1-\left \| u \right \|^{2})^{2}},
\end{equation}

From Eq. \ref{Dp_dist} and Eq. \ref{Def_X}, the following equation can be obtained,
\begin{equation}
\label{Dp_X}
d_{P}(u,v)=\cosh^{-1} (1+X).
\end{equation}

Expanding Eq. \ref{Dp_X} using a Taylor series and taking the first term yields
\begin{equation}
\label{Dp_X_E}
d_{P}(u,v)\approx \frac{2}{1-\left \| u \right \|^{2}}\left \| u-v \right \|.
\end{equation}

Thus, the ratio of the gradient magnitudes of $d_{P}$ and $d_{E}$ is:
\begin{equation}
\label{Dp_De}
\frac{\left \| \nabla _{u}d_{P} \right \|}{\left \| \nabla _{u}d_{E} \right \|} = \frac{2}{1-\left \| u \right \|^{2}} > 1.
\end{equation}

\end{proof}

This property provides a novel and powerful perspective for tackling complex image alignment challenges that are difficult to address within traditional Euclidean space. By facilitating more flexible representations and transformations, it opens up new possibilities for handling large deformations, non-linear distortions, and modality differences more effectively.

\subsection{Hyperbolic cycle Alignment Network (Hy-CycleAlign)}
The overall pipeline of our Hy-CycleAlign is shown in Fig. \ref{arc} (a). Taking infrared-to-visible alignment as an example, the network consists of two alignment stages during training. In the first stage, the original infrared image $T$ is aligned to the visible image $V$, producing the aligned image $T_{v}$. In the second stage, the aligned image $T_{v}$ is then inversely aligned back to the infrared image $T$, resulting in the reverse-aligned image $T_{vt}$. It is important to note that during the forward alignment process, the aligned result $T_{v}$ is also fused with $V$ to generate the final fusion image $F$.

Inspired by \cite{zhu2017unpaired}, we introduce an adversarial discriminator to assist the model in achieving better image registration performance. Specifically, we extract the gradient edges from from both the original infrared image $T$ and the aligned image $T_{v}$ as input to the discriminator, in order to distinguish the edges between the image $T$ and the aligned image $T_{v}$. Similarly, another discriminator is used to differentiate the edges between the reverse-aligned image $T_{vt}$ and the original image $T$. 

This design allows us to impose a consistency constraint between the reverse-aligned output and the input image.
Consequently, the model eliminates the need for additional high-precision, manually aligned image pairs, significantly reducing reliance on costly and time-consuming data annotation.

\subsection{Hyperbolic Hierarchy Contrastive Alignment (H$^{2}$CA)}
We design the H$^{2}$CA module to map images from Euclidean space to Poincar{\'e} space and apply constraints within the hyperbolic space. According to Eq. \ref{Project_E2P}, the vector $i\in \mathbb{R}^{d}$ in Euclidean space can be projected into Poincar{\'e} space using the function $Proj(*)$, where $c$ denotes the curvature.
\begin{equation}
\label{Project_E2P}
Proj(i)=\frac{i}{\sqrt{c}\cdot \left \| i \right \|} \tanh (\sqrt{c}\cdot \left \| i \right \|),
\end{equation}

For $m,n\in \mathbb{D}^{n}_{c}$ in the Poincar{\'e} space, M{\" o}bius addition is used in place of Euclidean addition, as shown in Eq. \ref{mobius}.
\begin{equation}
\label{mobius}
m\oplus n = \frac{(1+2c\left \langle m,n \right \rangle + c\left \| n \right \|^{2})m + (1-c\left \| m \right \|^{2})n}{1+2c\left \langle m,n \right \rangle+c^{2}\left \| m \right \|^{2}\left \| n \right \|^{2}} ,
\end{equation}

Thus, the distance between $m$ and $n$ in the Poincar{\'e} space can be calculated using Eq. \ref{pointcar_dis}.
\begin{equation}
\label{pointcar_dis}
d_{P}(m,n)=\frac{2}{\sqrt{c} } \cosh^{-1} (\sqrt{c} \left \| -m\oplus n \right \| ),
\end{equation}

According to Eq. \ref{pointcar_dis}, registration constraints can be implemented in the Poincar{\'e} space. It is important to note that the H$^{2}$CA imposes constraints on both pixels and edges separately, thus enabling the alignment of different hierarchies.

\subsection{Loss Function}
The Hy-CycleAlign model consists of four loss components: adversarial loss $\mathcal{L}_{adv}$, cycle consistency loss $\mathcal{L}_{cc}$, hyperbolic hierarchical contrastive alignment loss $\mathcal{L}_{h2c}$, and smoothness loss $\mathcal{L}_{sm}$.

\noindent \textbf{Adversarial loss $\mathcal{L}_{adv}$.}
We apply the adversarial loss to both registration networks. For the first registration $R_{t2v}(T,V): T \to V$ and its discriminator $D_{v}$, it can be formulated as Eq. \ref{L_adv}. It is noted that, 
$\mathcal{L}_{adv}$ constrains the edges of the image $V$ and the registered image $R_{t2v}(T,V)$, where $\nabla$ denotes the Sobel operator. Similarly, we use $\mathcal{L}_{adv}(V, T, R_{v2t}, D_{t})$ to constrain the alignment model $R_{v2t}(V,T): V \to T$ and the discriminator $D_{t}$ in the inverse alignment.
\begin{equation}
\label{L_adv}
\mathcal{L}_{adv}=\mathbb{E}_{v}[\log{D_{v}(\nabla V)}]+\mathbb{E}_{t}[\log{(1-D_{v}(\nabla R_{t2v}(T,V)))}].
\end{equation}

\noindent \textbf{Cycle consistency loss $\mathcal{L}_{cc}$.}
The input images $V$ and $T$ are aligned in two steps $R_{t2v}$ and $R_{v2t}$ to obtain the $V_{tv}$ and $T_{vt}$, respectively. Thus, we constrain the reconstructed images using the cyclic consistent loss as shown in Eq. \ref{L_cc}.
\begin{equation}
\label{L_cc}
\mathcal{L}_{cc}=\left \| T_{vt}-T \right \|_{1}+\left \| V_{tv}-V \right \|_{1}.
\end{equation}

\noindent \textbf{Hyperbolic hierarchical contrastive loss $\mathcal{L}_{h2c}$.}
We apply pixel-level contrastive constraints between $H(T_{v})$ and $H(V)$, where $H(*)$ denotes the hyperbolic mapping through the H$^{2}$CA module. Then, we impose structural-level contrastive constraints between the image edges $\nabla V$ and $\nabla T_{v}$, thereby achieving hierarchical constraints, as shown in Eq. \ref{L_hhc}. Similarly, we impose the same constraints on $T$ and $V_{t}$.
\begin{equation}
\label{L_hhc}
\mathcal{L}_{h2c}=-(\log{\sigma (-d_{P}(Tv,V))} + \log{\sigma (-d_{P}(\nabla Tv,\nabla V))}).
\end{equation}

\noindent \textbf{Smoothness loss $\mathcal{L}_{sm}$.}
To ensure the smoothness of the aligned image, we impose a constraint on the spatial gradient of the deformation field $\nabla f(\phi)$, as Eq. \ref{L_sm}.
\begin{equation}
\label{L_sm}
\mathcal{L}_{sm} = \sum_{\phi \in \Phi} \Vert \nabla f(\phi) \Vert ^2.
\end{equation}

\noindent \textbf{Fusion loss $\mathcal{L}_{f}$.}
Inspired by \cite{tang2022image, zhao2023cddfuse}, the fusion loss is defined as shown in Eq. \ref{L_fusion}, where $F$ denotes the fuse image with the size of $H\times W$.
\begin{equation}
\label{L_fusion}
\mathcal{L}_{f}=\frac{1}{HW}\left \| F-max(T_{v},V) \right \|_{1} + \frac{1}{HW} \left \| \left | \nabla F  \right | - max(\left | \nabla T_{v} \right | ,\left | \nabla V \right | ) \right \|_{1} .
\end{equation}

\noindent \textbf{Total loss $\mathcal{L}$.} 
Our total loss is:
\begin{equation}
\label{L_total}
\mathcal{L}=\mathcal{L}_{adv} + \mathcal{L}_{cc} + \mathcal{L}_{h2c} + \mathcal{L}_{sm} + \mathcal{L}_{f}.
\end{equation}

\section{Experiments}
\label{sec: experiment}

\begin{table*}[t]
  \setlength\tabcolsep{3pt}
  \centering
   \caption{Quantitative comparisons at DroneVehicle, LLVIP and MFNet. Note that the random nonlinear transformation is applied to both the infrared images in LLVIP and the visible images in MFNet. \textbf{Boldface} and \underline{underline} show the best and second-best values, respectively.}
   \label{result_all} 
  
 \resizebox{\linewidth}{!}{%
\begin{tabular}{ c c c c c c c c c c c}
    \toprule
    \multirow{2}*{ \textbf{\textcolor{black}{Dataset}}}& \multirow{2}*{\textbf{\textcolor{black}{Metric}}}
    
    & \multicolumn{3}{c}{\textbf{\textcolor{black}{Alignment-free fusion methods}}}
    & \multicolumn{6}{c}{\textbf{\textcolor{black}{Alignment-based fusion methods}}}\\

     \cmidrule(r){3-5} \cmidrule(r){6-11}   
   {}&{}& \textbf{ \textcolor{black}{DIDFuse} } & \textbf{\textcolor{black}{CDDFuse}} & \textbf{\textcolor{black}{EMMA}} & \textbf{ \textcolor{black}{SuperFusion} } & \textbf{\textcolor{black}{ReCoNet}} & \textbf{\textcolor{black}{MURF}} & \textbf{\textcolor{black}{UMF-CMGR}} & \textbf{\textcolor{black}{IMF}} & \textbf{\textcolor{black}{Hy-CycleAlign}}\\
    \cmidrule(r){1-11}

    \multirow{7}*{\textbf{\textcolor{black}{\rotatebox{90}{DroneVehicle}}}} 
    & \textbf{HD}$\downarrow$ & $\underline{74.20}$ & $76.41$ &$\mathbf{71.62}$ &$75.15$ & $73.91$ & $92.18$ &$\underline{65.27}$ & $\mathbf{65.19}$ & 	$70.36$\\
    {}&\textbf{HD95}$\downarrow$ &	$30.56$ &$\underline{29.11}$ & $\mathbf{27.23}$ &$\underline{30.12}$ &$30.53$ & $44.95$ & $31.73$& $30.97$&	$\mathbf{25.76}$\\
    {}&\textbf{ASSD}$\downarrow$ &	$8.17$	& $\underline{7.30}$ & $\mathbf{6.61}$ &$\underline{7.55}$&$7.71$& $12.06$ &	$8.93$&	$8.62$&	$\mathbf{6.38}$\\
    {}&\textbf{DSC}$\uparrow$ &	$\mathbf{0.80}$ &	$0.63$ &$\underline{0.72}$ &$\underline{0.67}$ &$\mathbf{0.75}$ & $0.66$ & $0.57$ & $0.59$&	$\mathbf{0.75}$\\
    {}&\textbf{MEE}$\downarrow$ & $39.37$ &	$\mathbf{31.59}$ & $\underline{32.12}$ &$\underline{28.55}$ & $32.84$ & $36.85$ &	 $33.56$& $34.71$ &	$\mathbf{28.28}$\\
    {}&\textbf{SF}$\uparrow$ &	$\underline{19.97}$ &$\mathbf{21.80}$ & $19.19$ &$\underline{17.50}$ &$13.60$ & $5.72$ & $10.96$&	$8.44$&	$\mathbf{21.45}$\\
    {}&\textbf{EN}$\uparrow$ & $6.95$ &$\mathbf{7.35}$ & $\underline{7.30}$ &$\underline{7.15}$ & $6.95$ & $6.83$ & $6.91$ &	 $7.07$ &	$\mathbf{7.18}$\\
    \cmidrule(r){1-11}

    \multirow{7}*{\textbf{\textcolor{black}{\rotatebox{90}{LLVIP}}}} 
    & \textbf{HD}$\downarrow$ & $203.33$& 	$\mathbf{159.21}$& 	$\underline{200.11}$&	$213.07$& 	$226.59$& $\underline{199.21}$ &	$234.82$&	$211.14$ & 	$\mathbf{163.49}$\\
    {}&\textbf{HD95}$\downarrow$& $104.27$ & $\mathbf{71.31}$& $\underline{71.57}$&	$120.91$&	$117.57$& $\mathbf{88.25}$ &	$140.75$&	$119.41$&	$\underline{113.50}$\\
    {}&\textbf{ASSD}$\downarrow$& $26.61$	& $\underline{17.19}$	& $\mathbf{17.16}$&	$29.97$&	$29.30$& $\underline{20.64}$ &	$39.17$&	$36.29$&	$\mathbf{15.50}$\\
    {}&\textbf{DSC}$\uparrow$&	$\mathbf{0.80}$ &	 $\underline{0.72}$& $0.71$&	$0.59$&	$0.48$& $\underline{0.79}$&$0.49$ &	$0.58$&	$\mathbf{0.85}$\\
    {}&\textbf{MEE}$\downarrow$& $33.78$ &	$\mathbf{18.19}$ & $\underline{18.47}$&	$\mathbf{18.08}$&	$33.42$& $19.05$ &	$23.11$&	$22.01$&	$\underline{18.30}$\\
    {}&\textbf{SF}$\uparrow$& $11.37$& $\mathbf{18.66}$ &$\underline{14.92}$&	$\underline{14.10}$&	$11.43$& $\mathbf{19.69}$ &	$4.64$&	$4.82$&	$11.27$\\
    {}&\textbf{EN}$\uparrow$&	$6.02$&	 $\mathbf{7.44}$& $\underline{7.36}$&	$\mathbf{7.34}$&	$5.85$&	$6.95$& $6.95$ &	$7.08$&	$\underline{7.19}$\\
    \cmidrule(r){1-11}

    \multirow{7}*{\textbf{\textcolor{black}{\rotatebox{90}{MFNet}}}} 
    & \textbf{HD}$\downarrow$ & 	$105.82$& 	$\underline{83.11}$& 	$\mathbf{72.14}$&	$108.63$& 	$110.36$& $\underline{76.77}$ &	$127.15$&	$87.87$ & 	$\mathbf{67.38}$\\
    {}&\textbf{HD95}$\downarrow$&	$58.32$&	$\mathbf{28.72}$&	$\underline{29.49}$&	$63.83$&	$69.72$& $\underline{30.47}$ &	$100.17$&	$68.03$&	$\mathbf{22.43}$\\
    {}&\textbf{ASSD}$\downarrow$&	$14.00$	&$\underline{6.75}$	&$\mathbf{6.44}$&	$14.49$&	$17.71$& $\underline{6.77}$ &	$39.25$&	$25.70$&	$\mathbf{4.43}$\\
    {}&\textbf{DSC}$\uparrow$&	$0.84$&	$\underline{0.90}$&	$\mathbf{0.94}$&	$0.84$&	$0.45$& $\underline{0.91}$&	$0.47$ &	$0.46$&	$\mathbf{0.93}$\\
    {}&\textbf{MEE}$\downarrow$&	$35.06$&	$\mathbf{7.83}$&	$\underline{10.08}$&	$12.53$&	$23.40$& $\mathbf{7.04}$ &	$16.74$&	$10.79$&	$\underline{7.39}$\\
    {}&\textbf{SF}$\uparrow$&	$8.50$&	$\mathbf{12.10}$&	$\underline{10.84}$&	$7.90$&	$9.51$& $\mathbf{10.47}$ &	$5.25$&	3.77&	$\underline{9.90}$\\
    {}&\textbf{EN}$\uparrow$&	$5.42$&	$\mathbf{6.58}$&	$\underline{6.57}$&	$6.21$&	$5.39$&	$\underline{6.38}$& $6.01$ &	$4.15$&	$\mathbf{6.50}$\\

\bottomrule
\end{tabular}
}
\end{table*}

\subsection{Setup}
\label{sec: setup}
\noindent \textbf{Datasets.}
We use the popular DroneVehicle \cite{sun2022drone}, LLVIP \cite{jia2021llvip} and MFNet \cite{ha2017mfnet} benchmarks to evaluate the performance of our model. Furthermore, to validate the effectiveness of different approaches in handling more complex misaligned multi-modal image fusion scenarios, we conduct experiments on the dataset, which is based on drone views. Given that the LLVIP and MFNet datasets have been manually aligned, it is necessary to construct misaligned images for both training and testing. To generate misaligned data, we apply random nonlinear transformations separately to the infrared images in the LLVIP dataset and the visible images in the MFNet dataset.

\noindent \textbf{Metrics.}
We use six metrics to quantitatively measure the alignment and fusion results of the model: hausdorff distance (HD), $95\%$ hausdorff distance (HD95), average symmetric surface distance (ASSD), dice similarity coefficients (DSC), median square error (MEE), spatial frequency (SF), and entropy (EN). 
These metrics provide a comprehensive evaluation of aligned and fused image quality, detail retention, information integrity, and visual perception performance.

\noindent \textbf{Implement details.}
Hy-CycleAlign needs to be trained for 120 epochs. All network parameters are updated with the AdamW optimizer \cite{loshchilov2017fixing} with the initial learning rate set to $10^{-4}$. The effective edge threshold $c$ is $0.01$.

\subsection{Comparing with SOTA}
\label{compar}

In this sction, we test Hy-CycleAlign on the three test sets and compare the alignment and fusion results withe the state-of-the-art methods including DIDFuse \cite{zhao2020didfuse}, CDDFuse \cite{zhao2023cddfuse}, EMMA \cite{Zhao_2024_CVPR}, SuperFusion \cite{TANG2022SuperFusion_}, ReCoNet \cite{huang2022reconet}, MURF \cite{xu2023murf}, UMF-CMGR \cite{wang2022unsupervised}, IMF \cite{wang2024improving}. SuperFusion, ReCoNet, MURF, UMF-CMGR, and IMF are currently the mainstream alignment-based fusion methods.

\begin{figure}[t]
\centering
\includegraphics[width=1.\columnwidth]{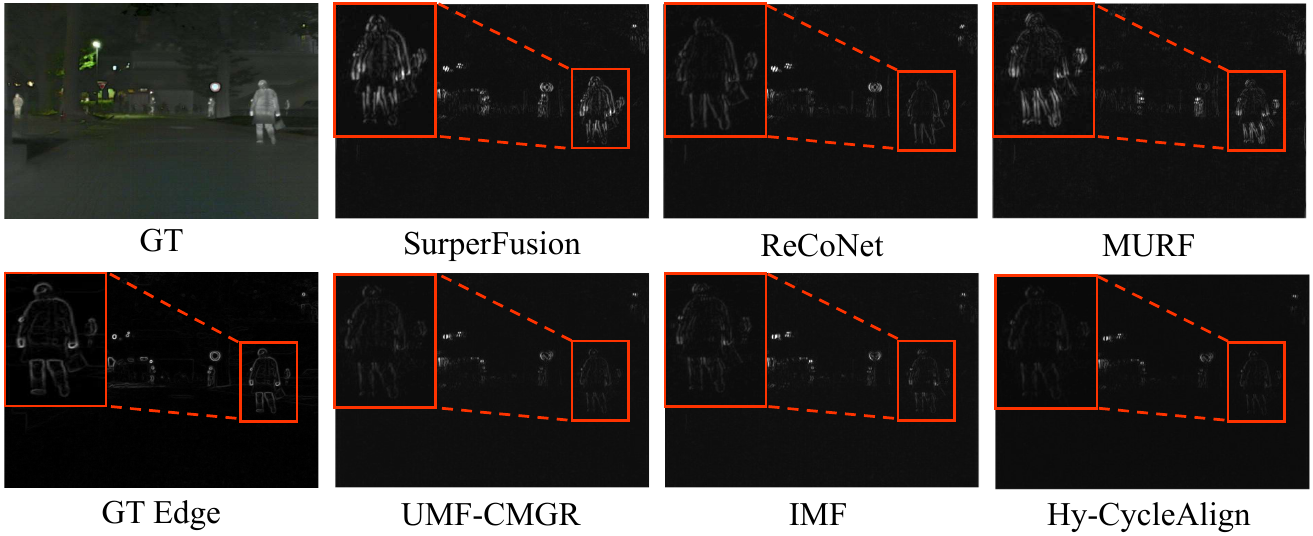} 
\caption{The aligned MFNet dataset is used as ground truth (GT) to compare the edge intensity differences after applying the alignment method.}
\label{result_diff}
\end{figure}

\noindent \textbf{Qualitative comparison of alignment effects.} 
We use the fusion results obtained by training and testing EMMA on data without deformation as the ground truth for registration and fusion methods. Edge features are extracted using the Sobel operator and compared with those from the fusion results of existing registration-based methods. Obviously, our method gives maintains good alignment results when facing targets with significant edge differences, the intensity differences are shown in Fig. \ref{result_diff}.

\begin{figure}[t]
\centering
\includegraphics[width=1.\columnwidth]{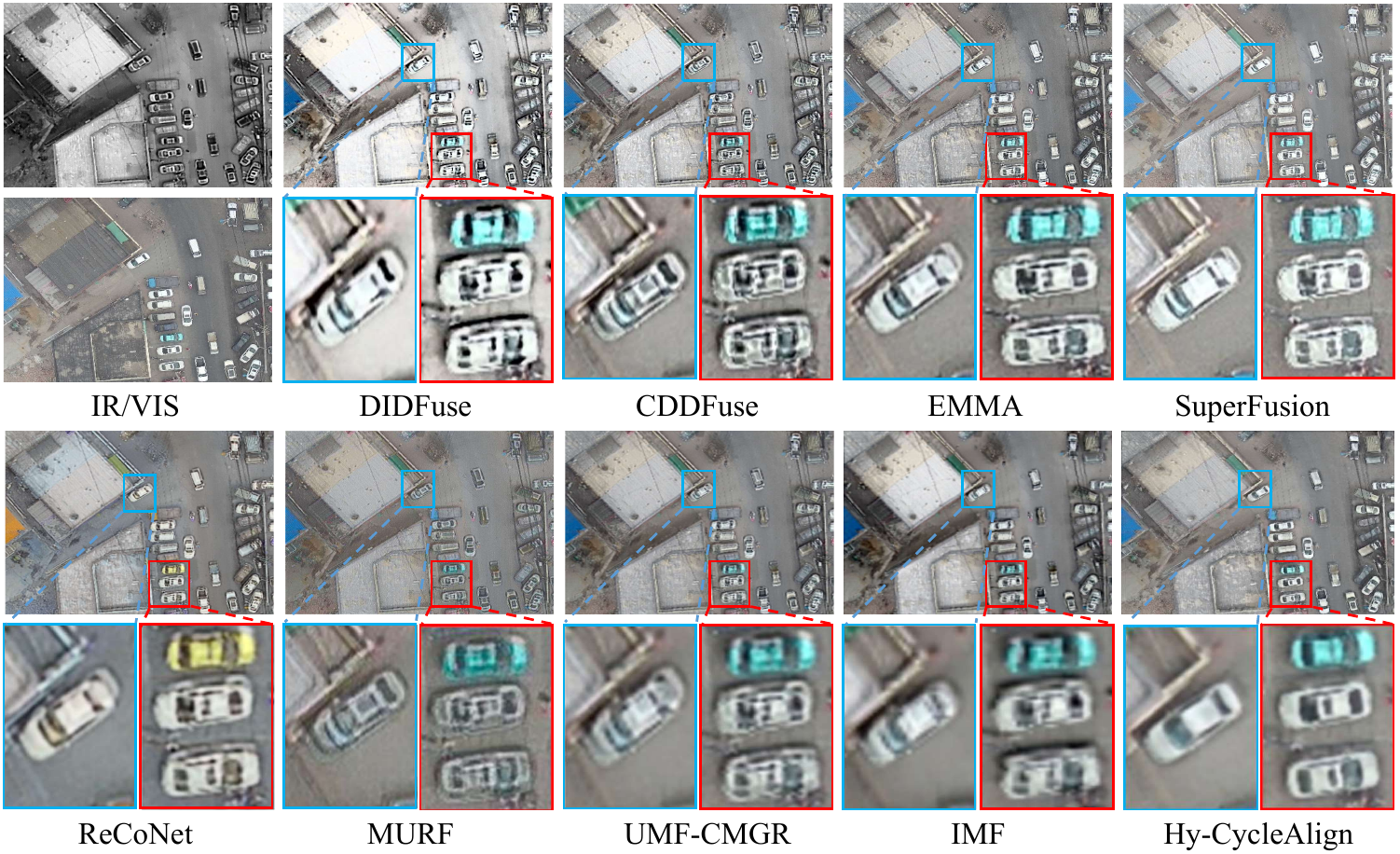} 
\caption{Comparison of results in a DroneVehicle dataset based on drone views.}
\label{result_dv}
\end{figure}

\begin{figure}[t]
\centering
\includegraphics[width=1.\columnwidth]{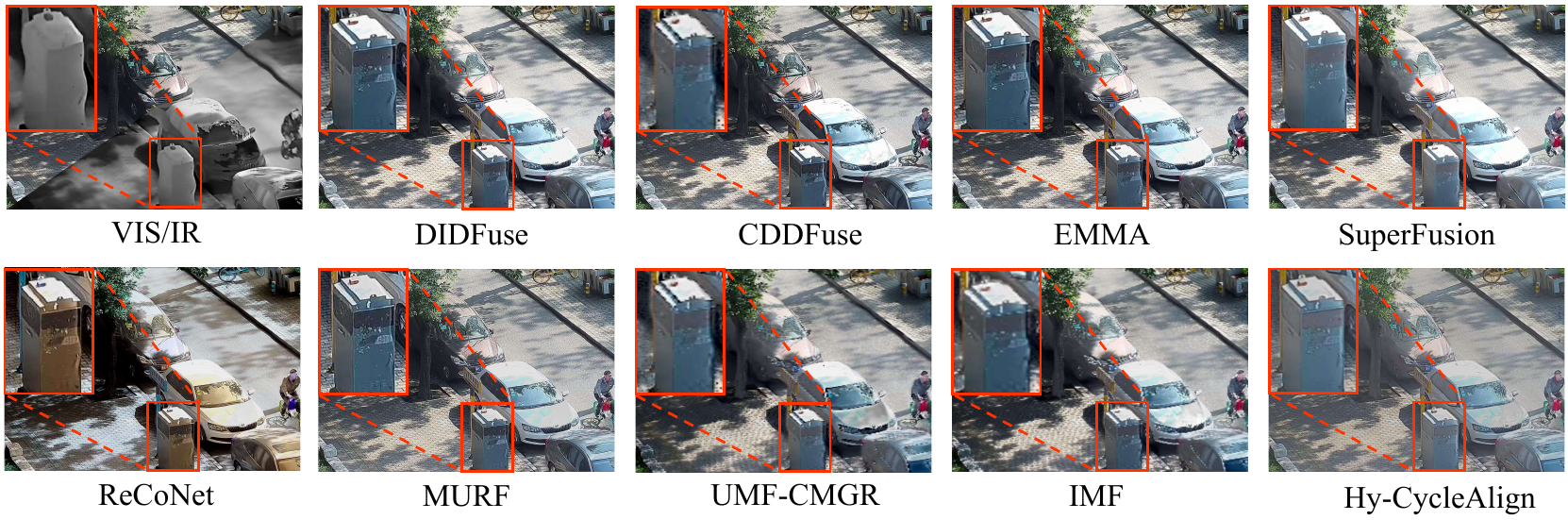} 
\caption{Comparison of results for the LLVIP dataset with IR nonlinear transformations.}
\label{result_llvip}
\end{figure}

\begin{figure}[t]
\centering
\includegraphics[width=1.\columnwidth]{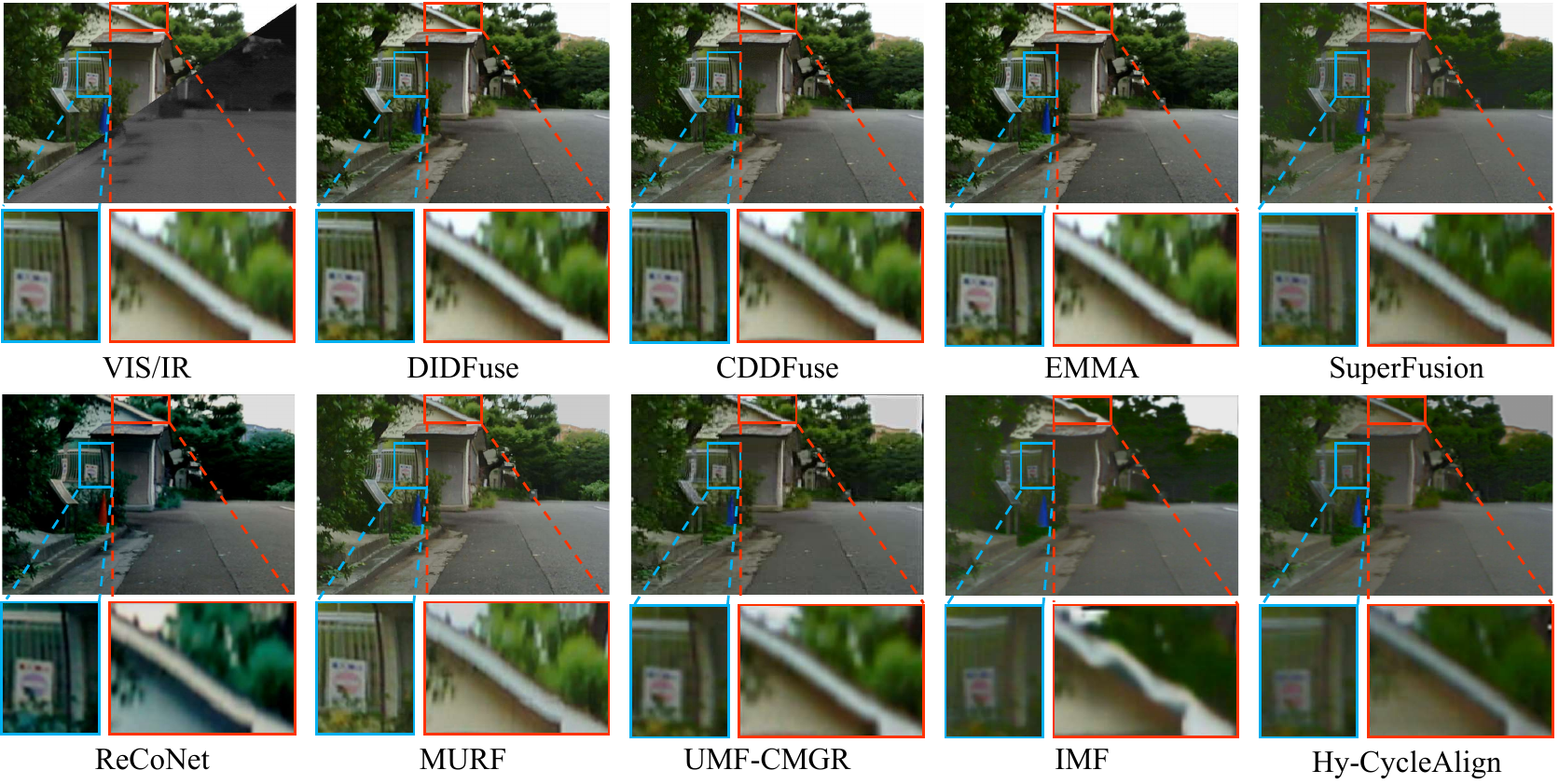} 
\caption{Comparison of results for the MFNet dataset with visible nonlinear transformations.}
\label{result_mfnet}
\end{figure}

\noindent \textbf{Qualitative comparison of fusion effects.} 
Fig. \ref{result_dv}, \ref{result_llvip} and \ref{result_mfnet} show the fusion results under different misalignment conditions. It is clear that our method achieves robust alignment and fusion performance in various types of misalignment and in diverse scenes. Compared with existing methods, our Hy-CycleAlign achieves better registration performance, and no incorrect registration results are observed.

\noindent \textbf{Quantitative comparison.} 
We conducted a quantitative comparison using five commonly used registration metrics and two fusion metrics, as shown in Table \ref{result_all}. Our method achieved the best registration and fusion performance on the real-world misaligned dataset DroneVehicle. Similarly, it performed excellently on the nonlinearly misaligned MFNet dataset. Although the performance advantage on the LLVIP dataset is less pronounced compared to the other two datasets, overall, our method still demonstrates strong overall competitiveness.

\subsection{Ablation studies}
\begin{figure}[h]
\centering
\begin{minipage}[t]{0.45\textwidth}
\centering
\includegraphics[width=\textwidth]{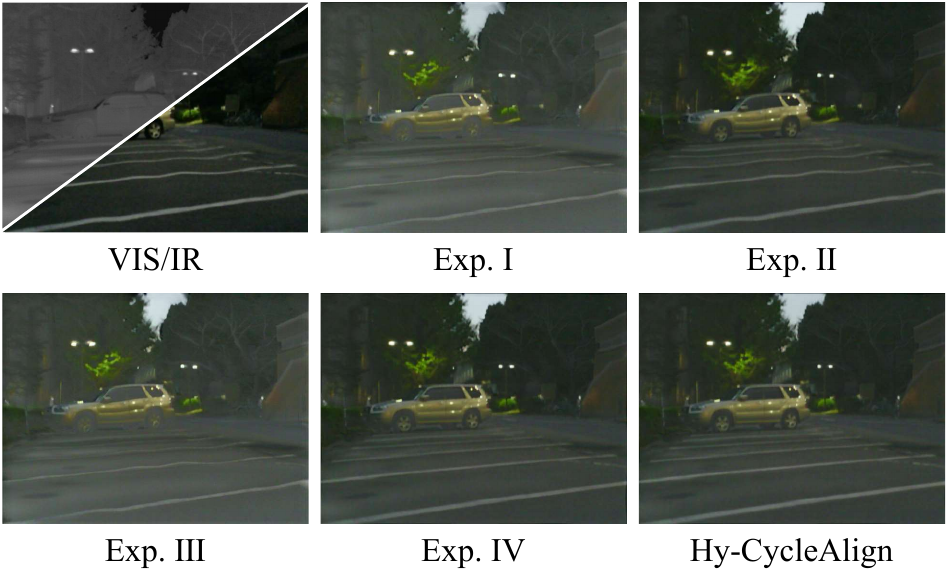} 
\captionof{figure}{The analysis of the ablation experiment was conducted using the MFNet dataset.}
\label{result_ab}
\end{minipage}
\hfill
\begin{minipage}[b]{0.49\textwidth}
\centering

\captionof{table}{Ablation experiment results in the test set of MFNet. CA denotes whether a cycle alignment structure, Eu refers to alignment in Euclidean space, $H^{2}CA\text{-}p$ denotes pixels alignment in hyperbolic space, and $H^{2}CA\text{-}e$ denotes edge alignment in hyperbolic space. Bold indicates the best values.}
\label{table_abe}
\resizebox{\textwidth}{!}{%
\begin{tabular}{c|c c c c|c c c c c}

\toprule
Methods &$Eu$ &$CA$ &$H^{2}CA\text{-}p$ &$H^{2}CA\text{-}e$ &HD $\downarrow$ & HD95 $\downarrow$ & ASSD $\downarrow$ & DSC $\uparrow$ & EN $\uparrow$ \\
\midrule
Exp. \uppercase\expandafter{\romannumeral1} &\Checkmark &\Checkmark &  & &$153.51$ &$86.68$  &$22.99$ &$0.72$ &$6.43$ \\
Exp. \uppercase\expandafter{\romannumeral2} & & &\Checkmark &\Checkmark &$96.13$&$38.44$ &$8.91$  &$0.92$ &$6.43$ \\
Exp. \uppercase\expandafter{\romannumeral3} & &\Checkmark &\Checkmark & &$95.76$&$36.15$ &$8.47$  &$0.93$ &$6.42$ \\
Exp. \uppercase\expandafter{\romannumeral4} & &\Checkmark & &\Checkmark &$93.91$&$34.71$ &$7.88$  &$0.92$ &$6.44$ \\
\midrule
Hy-CycleAlign &  &\Checkmark &\Checkmark &\Checkmark &$\mathbf{67.38}$ &$\mathbf{22.43}$ &$\mathbf{4.43}$ &$\mathbf{0.93}$ &$\mathbf{6.50}$ \\
\bottomrule

\end{tabular}
}
\end{minipage}
\end{figure}

Ablation experiments are set to verify the rationality of the different modules. HD, HD95, ASSD, DSC and EN. The results of experimental groups are shown in Fig. \ref{result_ab} and Tab. \ref{table_abe}.

\noindent \textbf{Euclidean space alignment baseline (Eu).}  In Exp. \uppercase\expandafter{\romannumeral1}, we retained the cyclic adversarial registration network structure and used the Sobel operator to extract edge features from the registered images, applying constraints in Euclidean space. The experimental results indicate that it is difficult to achieve multi-modal image registration in Euclidean space.

\noindent \textbf{Cycle consistent alignment structure (CA).} In Exp. \uppercase\expandafter{\romannumeral2}, we removed the reverse alignment process to verify the role of the cyclic alignment structure in the registration task. The results show that the alignment performance improved across all evaluation metrics.

\noindent \textbf{Pixel alignment ($H^{2}CA\text{-}p$) and edge alignment ($H^{2}CA\text{-}e$) in hyperbolic space.}  In Exp. \uppercase\expandafter{\romannumeral3} and Exp. \uppercase\expandafter{\romannumeral4}, we evaluate pixel-level alignment and edge alignment in hyperbolic space, respectively. Both approaches independently improve registration performance, but when combined, they complement each other and further enhance the overall alignment effectiveness.

\begin{figure}[t]
\centering
\includegraphics[width=1.\columnwidth]{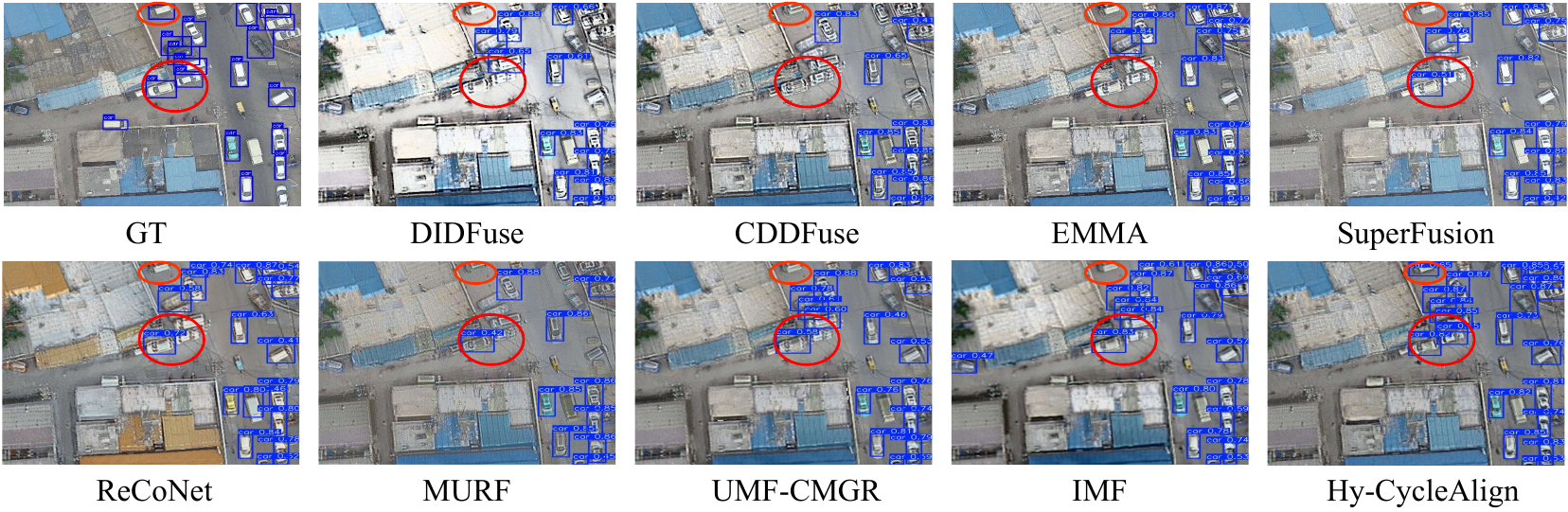} 
\caption{Comparison of fusion and detection results on misaligned data from the DroneVehicle.}
\label{result_det}
\end{figure}

\subsection{Downstream applying alignment and fusion}
To further show that the alignment task can  effectively enhance fusion and its downstream tasks, we apply the methods compared in Section \ref{compar} to an object detection task. The experimental results are shown in Fig. \ref{result_det}. Due to space limitations, more experimental results and analyses will be provided in the supplementary material.

\subsection{Additional Analysis}
We compared the computational complexity (FLOPs) and the number of parameters between our model and existing registration models. The results are shown below. Although the transformation from Euclidean space to hyperbolic space introduces more parameters, it does not lead to a significant increase in computational complexity.

\begin{table}[h]
\centering
\caption{Comparison of network parameter quantities.}
\label{result_det}
\begin{tabular}{c|cccccc}
\hline
 & SuperFusion & ReCoNet & MURF & UMF & IMF & Ours \\ \hline
FLOPs (G) & 16.36 & 15.33 & 100.10 & 131.38 & 123.30 & 16.66 \\ 
Parameters (M) & 1.96 & 0.21 & 4.08 & 14.44 & 15.70 & 18.26 \\ \hline
\end{tabular}
\end{table}

\subsection{Limitation}
\label{sec: discussion}

Although our proposed Hy-CycleAlign is the first method to perform multi-modal image registration in hyperbolic space and demonstrates promising results in infrared-visible alignment tasks, several limitations still remain.  Future work should focus on improving the computational efficiency of operations within Poincar{\'e} space and enhancing the model's adaptability to images with large modality discrepancies, particularly in real-time or large-scale scenarios.  Given the challenges posed by the current Poincar{\'e} model in handling complex alignment cases, exploring more advanced or hybrid hyperbolic geometries, combined with adaptive embedding mechanisms and constraint strategies, represents a crucial direction for improving the generality and robustness of hyperbolic registration frameworks.


\section{Conclusions}
This paper proposes a hyperbolic cyclic alignment and fusion model. By leveraging forward and backward alignment constraints in a cyclic manner, the model effectively performs multi-modal image registration. Hy-CycleAlign is the first registration framework based on hyperbolic space. It maps images from Euclidean space to hyperbolic space and imposes multi-level alignment constraints, which alleviates modality discrepancies. Finally, we provide corresponding theoretical proofs. Experimental results demonstrate that Hy-CycleAlign achieves promising performance in infrared and visible image alignment and fusion.

\appendix
\section{More Explanations of the Motivation}
Infrared and visible images have modal differences due to differences in imaging principles, making them nonlinear \cite{jiang2023breaking}. Although Euclidean space handles linear discrepancies well, its inherently linear geometry limits its ability to represent complex nonlinear relationships \cite{ganea2018hyperbolic}. Therefore, it struggles to accurately capture the nonlinear modality gaps, making it inadequate to address cross-modal discrepancies in such settings. Due to the negative curvature of the hyperbolic space, it is naturally good at modeling nonlinear relationships \cite{ramasinghe2024accept, zhang2023multimodal}. We explore multi-modal image registration within hyperbolic space. By leveraging its powerful capacity to represent complex structures, hyperbolic space enables more accurate capture of the nonlinear differences between modalities.



\begin{theorem}
Compared to Euclidean space, the hyperbolic space represented by the Poincar{\'e} space is more sensitive to misalignments, and this sensitivity increases as points approach the boundary of the Poincar{\'e} space.
\end{theorem}

\begin{proof}
Assuming $u$ and $v$ are the points to be registered from different modality images, their distance in Euclidean space $d_{E}(u,v)$ can be expressed as
\begin{equation}
\label{De_dist}
d_{E}(u,v)=\left \| u-v \right \| _{2}.
\end{equation}

Assuming the normal Poincar{\'e} space $\mathbb{D}^{n}=\{ x\in \mathbb{R}^{n}:\left \| x \right \| < 1 \}$, $x$ denotes a point in the Poincar{\'e} space. Then, the distance $d_{p}(u,v)$ between points u and v in the Poincar{\'e} space is shown as
\begin{equation}
\label{Dp_dist}
d_{P}(u,v)=\cosh^{-1} (1+2\frac{\left \| u-v \right \|^{2}}{(1-\left \| u \right \|^{2})(1-\left \| v \right \|^{2})} ).
\end{equation}

Let $\delta =v-u$, in the alignment task, the goal is to make $v\to u$. Then, it follows that $\left \| v \right \| \approx  \left \| u \right \| $.

It can be further shown that
\begin{equation}
\label{Dp_dist}
1-\left \| u \right \|^{2} \approx 1-\left \| v \right \|^{2}.
\end{equation}

We define X in Eq. \ref{Def_X}, 
\begin{equation}
\label{Def_X}
X=2\frac{\left \| u-v \right \|^{2}}{(1-\left \| u \right \|^{2})(1-\left \| v \right \|^{2})}\approx 2\frac{\left \| \delta  \right \|^{2}}{(1-\left \| u \right \|^{2})^{2}},
\end{equation}

Substituting Eq. \ref{Def_X} into Eq. \ref{Dp_dist}, the distance computation in the Poincar{\'e} space becomes:
\begin{equation}
\label{Dp_X}
d_{P}(u,v)=\cosh^{-1} (1+X).
\end{equation}

According to the Taylor series expansion, we obtain:
\begin{equation}
\label{Taylor_total}
\cosh^{-1} (1+X) = \sqrt{2X}+o(X^{3/2}).
\end{equation}

Neglecting the minimum term of the above equation, we know that
\begin{equation}
\label{Taylor_negmin}
\cosh^{-1} (1+X) \approx \sqrt{2X}.
\end{equation}

Expanding Eq. \ref{Dp_X} using a Taylor series and taking the first term yields
\begin{equation}
\label{Dp_X_E}
d_{P}(u,v)\approx \frac{2}{1-\left \| u \right \|^{2}}\left \| u-v \right \|.
\end{equation}

Thus, the ratio of the gradient magnitudes of $d_{P}$ and $d_{E}$ is:
\begin{equation}
\label{Dp_De}
\frac{\left \| \nabla _{u}d_{P} \right \|}{\left \| \nabla _{u}d_{E} \right \|} = \frac{2}{1-\left \| u \right \|^{2}} > 1.
\end{equation}

Moreover, as $u$ approaches 1, it follows that $1-\left \| u \right \| ^{2} \to 0$ and in this case,
\begin{equation}
\label{Dp_dp_u_2_1}
\left \| \nabla _{u}d_{P} \right \| \to \infty.
\end{equation}

\end{proof}

Through this process, we demonstrate that image registration in hyperbolic space is more sensitive than in Euclidean space. This indicates that even slight misalignments lead to more significant changes in hyperbolic space, suggesting that applying registration constraints in hyperbolic space theoretically yields better results. Moreover, the closer the mapped pixels are to the boundary of the Poincar{\'e} space, the more sensitive they become to minor misalignments.


\section{More details of Hy-CycleAlign}
Hy-CycleAlign simultaneously trains two registration networks, $R_{t2v}$: $T$ to $V$ and $R_{v2t}$: $V$ to $T$, two discriminators, $D_{t}$ and $D_{v}$, and a fusion network. $R_{t2v}$ aligns the infrared image to the visible image, producing the registered image $T_{v}$. $R_{v2t}$ aligns the visible image to the infrared image, generating the registered image $V_{t}$. The fusion network then fuses $T_{v}$ and $V$ to generate the final fused image $F$.

We provide implementation details of the training phase of Hy-CycleAlign to clearly describe the training process, as shown in Algorithm \ref{alg1}.

\begin{algorithm}[t]

\caption{Pseudocode for Hy-CycleAlign training phase.}
\KwIn{\rm unaligned \ multi-modal \ images \ \it V \ and \ T \\}
\KwOut{\rm fusion \ images \it F \\ }

\BlankLine
\For{\ {V, T} \ \textbf{in} \ Dataloader \ }{

$\phi_{t2v} = R_{t2v}(T,V)$ \hfill // Generate the deformation field for infrared-to-visible alignment

$\phi_{v2t} = R_{v2t}(T,V)$ \hfill // Generate the deformation field for visible-to-infrared alignment

$T_{v} = T \circ \phi_{t2v}, V_{t} = V \circ \phi_{v2t} $ \hfill // Generate aligned images

$T_{vt} = T_{v} \circ \phi_{v2t}, V_{tv} = V_{t} \circ \phi_{t2v} $ \hfill // Generate reverse-aligned images

$H^{2}CA(\nabla T_{v},\nabla V), H^{2}CA(\nabla V_{t},\nabla T) $ \hfill // $H^{2}CA-e:$ Edge-to-Poincar{\'e} embedding 

$H^{2}CA(T_{v},V), H^{2}CA(V_{t},T) $ \hfill // $H^{2}CA-p:$ Pixel-to-Poincar{\'e} embedding

$D_{v}(\nabla T_{v},\nabla V), D_{t}(\nabla T,\nabla V_{t})$ \hfill // Determining alignment performance

$F =\rm Decoder(\rm Encoder(\it V \rm) + Encoder(\it T_{v} \rm)\rm )$ \hfill // Fusion of aligned multi-modal images

\label{alg1}
}

\end{algorithm}

\subsection{More Details of Architecture}
Hy-CycleAlign simultaneously trains two registration networks, $R_{t2v}$: $T$ to $V$ and $R_{v2t}$: $V$ to $T$, two discriminators, $D_{t}$ and $D_{v}$, and a fusion network. $R_{t2v}$ aligns the infrared image to the visible image, producing the registered image $T_{v}$. $R_{v2t}$ aligns the visible image to the infrared image, generating the registered image $V_{t}$. 
The fusion network then fuses $T_{v}$ and $V$ to generate the final fused image $F$.

\subsection{More Details of H$^{2}$CA}
H$^{2}$CA consists of two parts, H$^{2}$CA-e andH$^{2}$CA-p, which map image pixels and edge information from Euclidean space to the Poincar{\'e} space.

\noindent \textbf{H$^{2}$CA-e:}
H$^{2}$CA-e is used to map image edge information from Euclidean space to the Poincar{\'e} space. Edge features are first extracted using the Sobel operator $\nabla$. Inspired by \cite{khrulkov2020hyperbolic}, the extracted edge information is projected onto the hyperbolic tangent space, as shown in Eq. \ref{expmap0}.
\begin{equation}
\label{expmap0}
x=Map(i,c)=\frac{i}{\sqrt{c}\cdot \left \| i \right \|} \tanh (\sqrt{c}\cdot \left \| i \right \|),
\end{equation}
where the vector $i\in \mathbb{R}^{d}$ in Euclidean space can be projected into Poincar{\'e} space using the function $Map(*)$, where $c$ denotes the curvature.

Then, to avoid $x \ge \frac{1}{\sqrt{c}}$, i.e., to prevent the mapped values from exceeding the boundary of the Poincar{\'e} space, we use Equation \ref{project} to ensure that the projection lies within the Poincar{\'e} space.
\begin{equation}
\label{project}
Proj(x,c) = \begin{cases}
 x & \text{ if } \left \| x \right \|<\frac{1}{\sqrt{c}}-\varepsilon  \\
 \frac{(1-\varepsilon)}{\sqrt{c}}\frac{x}{\left \| x \right \|}   & \text{otherwise}
\end{cases}.
\end{equation}

To prevent overflow beyond the boundary, we introduce a small constant $\varepsilon$ and set it to $10^{-6}$.

We constrain image alignment by computing the geodesic distance between different modalities in hyperbolic space and converting it into a similarity probability:
\begin{equation}
\label{h2c-e}
\mathcal{L}_{h2c-e}=-\log{\sigma (-d_{P}(\nabla Tv,\nabla V))}.
\end{equation}

\noindent \textbf{H$^{2}$CA-p:}
Different from H$^{2}$CA-p, which constrains edge information, H$^{2}$CA-p focuses on aligning deep features. It first extracts features from each modality using a VGG-16 network \cite{simonyan2014very}, then maps them into the Poincar{\'e} space to achieve global multi-modal alignment:
\begin{equation}
\label{h2cp}
\mathcal{L}_{h2c-p}=-(\log{\sigma (-d_{P}(Tv,V))}.
\end{equation}

Therefore, the total loss in H$^{2}$CA is:
\begin{equation}
\label{L_hhc}
\mathcal{L}_{h2c}=\mathcal{L}_{h2c-e} + \mathcal{L}_{h2c-p}.
\end{equation}

\subsection{More Details of Fusion Module}
To eliminate the influence of complex fusion strategies on experimental conclusions, we deliberately adopt a simple fusion architecture to more clearly verify the direct relationship between registration quality and final fusion performance. The fusion module uses two encoders to extract features from the registered infrared and visible images, and then fuses them to produce the final fused output.

\section{More Experiments}
To validate the performance of Hy-CycleAlign, we conducted additional experiments with different negative curvatures $c$ in Poincar{\'e} space and tested the model on various misaligned data. The results demonstrate that Hy-CycleAlign consistently achieves good registration and fusion performance, even under different misalignment conditions.
\subsection{Hyperparametric Analysis}
\begin{figure}[t]
\centering
\begin{minipage}[h]{0.45\textwidth}
\centering
\includegraphics[width=\textwidth]{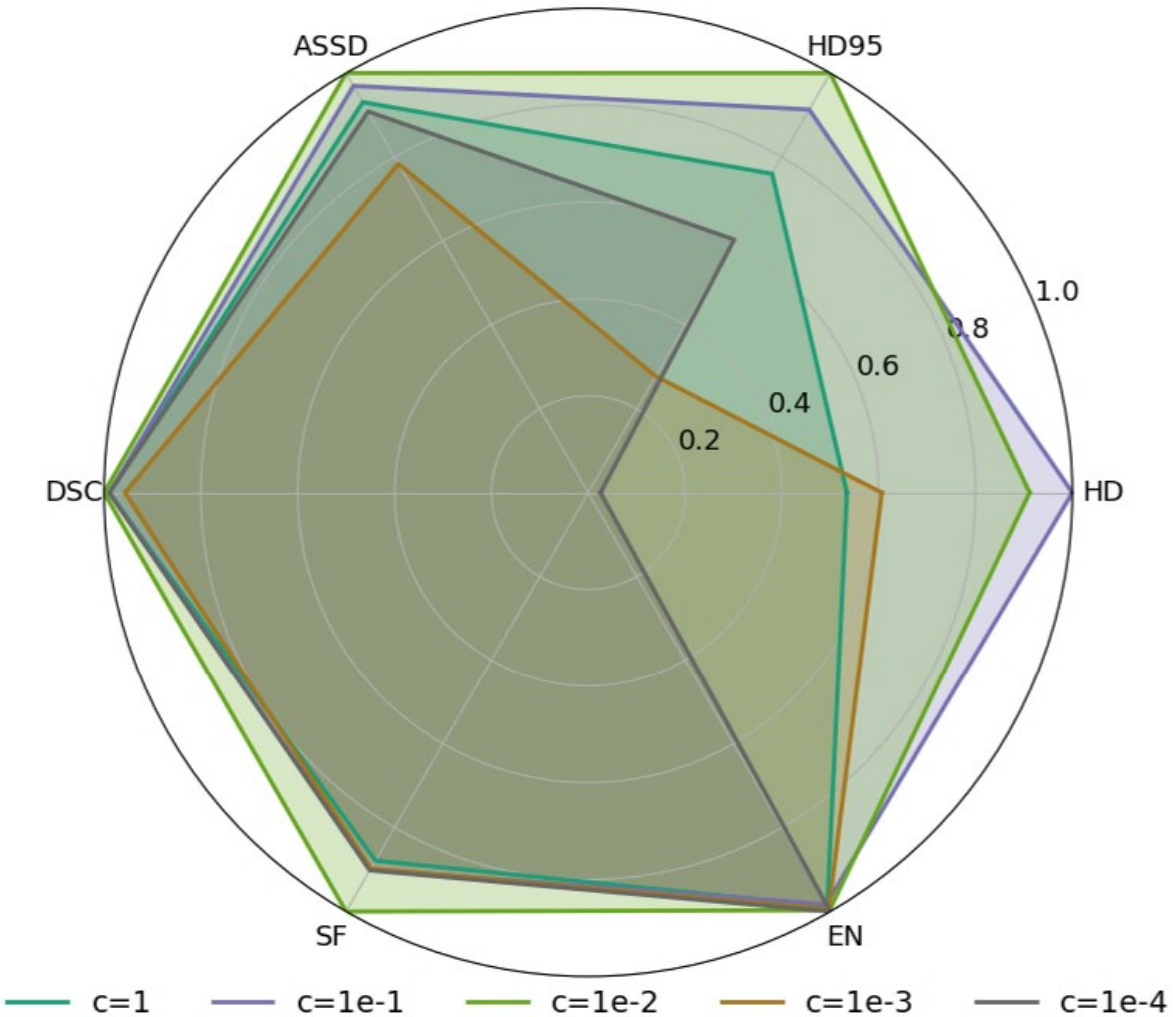} 
\captionof{figure}{Comparison of results in MFNet dataset for hyperparameter $c$.}
\label{result_abc}
\end{minipage}
\hfill
\begin{minipage}[h]{0.54\textwidth}
\centering

\captionof{table}{Ablation study results of the hyperparameter $c$ on the MFNet dataset. Bold indicates the best value.}
\label{table_abe}
\resizebox{\textwidth}{!}{%
\begin{tabular}{c|c c c c c}

\toprule
$c$ & HD $\downarrow$ & HD95 $\downarrow$ & ASSD $\downarrow$ & DSC $\uparrow$ & EN $\uparrow$ \\
\midrule
$1$ &$96.48$&$38.38$ &$8.72$  &$0.91$ &$6.38$ \\
$1e-1$ &$\mathbf{93.32}$&$36.04$ &$8.25$  &$0.91$ &$6.35$ \\
$1e-2$ &$93.91$&$\mathbf{34.71}$ &$\mathbf{7.88}$  &$\mathbf{0.92}$ &$6.44$ \\
$1e-3$ &$95.95$&$45.76$ &$10.50$  &$0.88$ &$6.42$ \\
$1e-4$ &$99.83$&$40.78$ &$8.99$  &$0.91$ &$\mathbf{6.46}$ \\
\bottomrule

\end{tabular}
}
\end{minipage}
\end{figure}

We analyzed the negative curvature $c$ of the Poincar{\'e} space. The experimental results are shown in Fig. \ref{result_abc} and Tab. \ref{table_abe}. Hy-CycleAlign achieves good visual registration results across different values of the parameter $c$. Combined with quantitative results, setting $c$ to $0.01$ yields a balanced performance in both registration and fusion tasks.

\begin{figure}[t]
  \centering
  \includegraphics[width=1\linewidth]{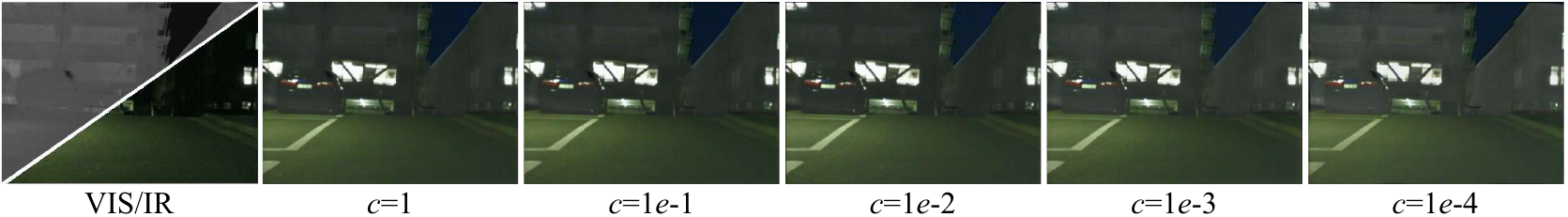}
   \caption{Visualization of the MFNet dataset with hyperparameter $c$.}
   \label{result_rs_sup}
\end{figure}

\subsection{More Downstream Results for Infrared-visible Applications}
\begin{figure}[t]
  \centering
  \includegraphics[width=1\linewidth]{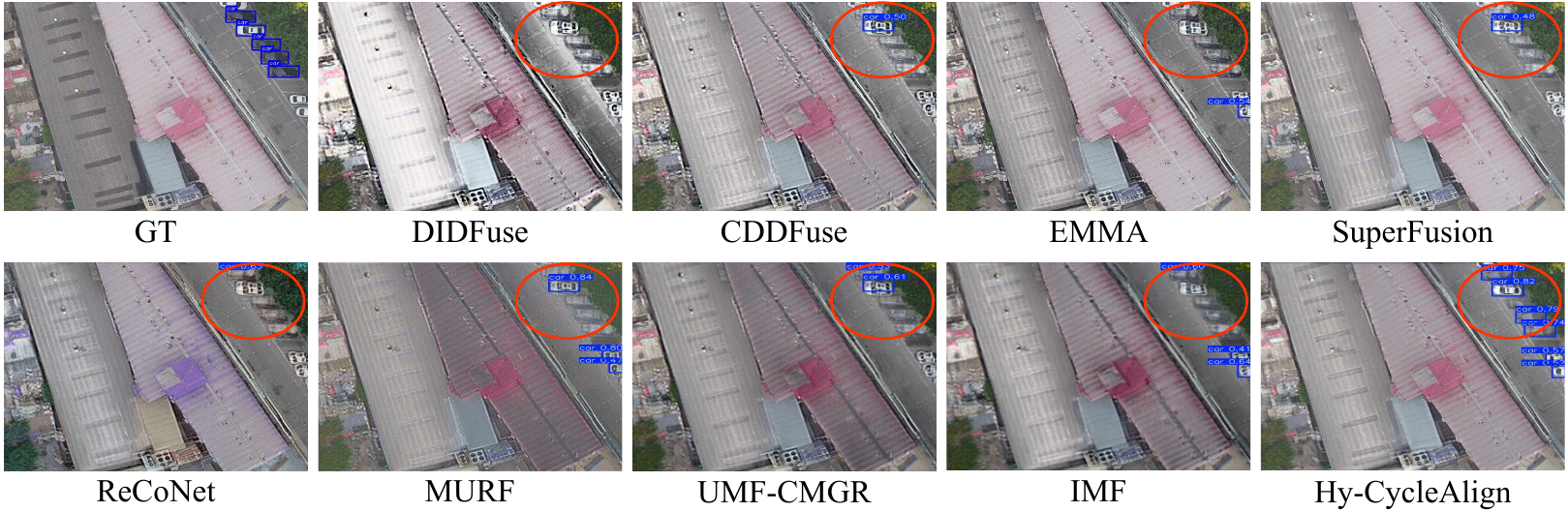}
   \caption{Qualitative results for infrared-visible object detection on DroneVehicle dataset}
   \label{result_det2}
\end{figure}

We apply the registration and fusion results to the task of object detection from the viewpoint of drones. In this task, we use YOLOv11-m \cite{yolo11_ultralytics} as the detector and employ mAP@0.5, precision and recall as evaluation metrics. Compared with existing methods, Hy-CycleAlign achieves the highest detection precision and mAP, indicating that registration and fusion can effectively enhance the performance of object detection tasks. The results are shown in Fig. \ref{result_det2} and Tab. \ref{table_det2}.

\begin{table}[t]
\centering
\caption{Quantitative results of object detection in the DroneVehicle dataset. Bold and underline indicate the best and second-best values, respectively.}
\label{table_det2}
\resizebox{\columnwidth}{!}{
\begin{tabular}{c|c c c c c c c c c}
\toprule
Methods & DIDFuse & CDDFuse & EMMA & SuperFusion & ReCoNet & MURF & UMF-CMGR & IMF & Hy-CycleAlign \\
\midrule
Recall $\uparrow$ &$73.8$ &$80.0$  &$70.8$ &$69.2$ &$78.5$ &$53.8$ &$80.8$ &$\mathbf{83.1}$ &$\underline{81.5}$\\
Precision $\uparrow$ &$7.7$&$7.0$ &$8.7$ &$8.3$ &$\underline{8.9}$ &$7.0$ &$\underline{8.9}$ &$8.3$ &$\mathbf{12.3}$ \\
mAP@0.5 $\uparrow$ &$7.2$&$6.5$ &$\underline{8.7}$ &$5.3$ &$8.5$ &$7.3$ &$6.9$ &$8.2$ &$\mathbf{13.7}$ \\
\bottomrule
\end{tabular}
}
\end{table}

\subsection{Comparisons of Rigid Misalignment Fusion Results}
\begin{figure}[t]
  \centering
  \includegraphics[width=1\linewidth]{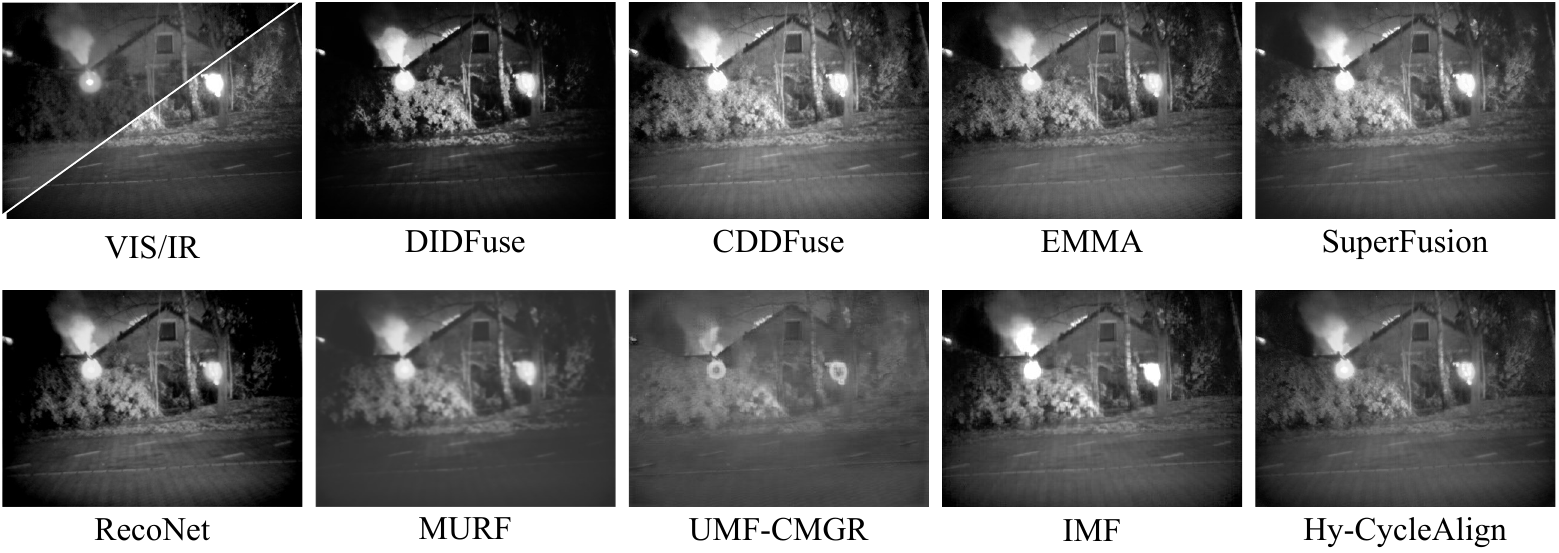}
   \caption{Comparison of results for the TNO dataset.}
   \label{result_tno_sup}
\end{figure}

To further validate the alignment and fusion performance of Hy-CycleAlign, we conducted additional experiments on the RoadScene \cite{xu2020aaai} and TNO \cite{toet2012progress} datasets. Considering that RoadScene is a well-aligned dataset, we randomly shifted the infrared images horizontally by 0.5\% to 1.5\% of the image width to artificially generate rigid misalignments caused by camera position differences. As shown in Fig. \ref{result_rs_sup} and \ref{result_tno_sup}, under varying lighting conditions, Hy-CycleAlign maintains good alignment performance when facing rigid misalignment while better preserving image details.

\begin{figure}[t]
  \centering
  \includegraphics[width=1\linewidth]{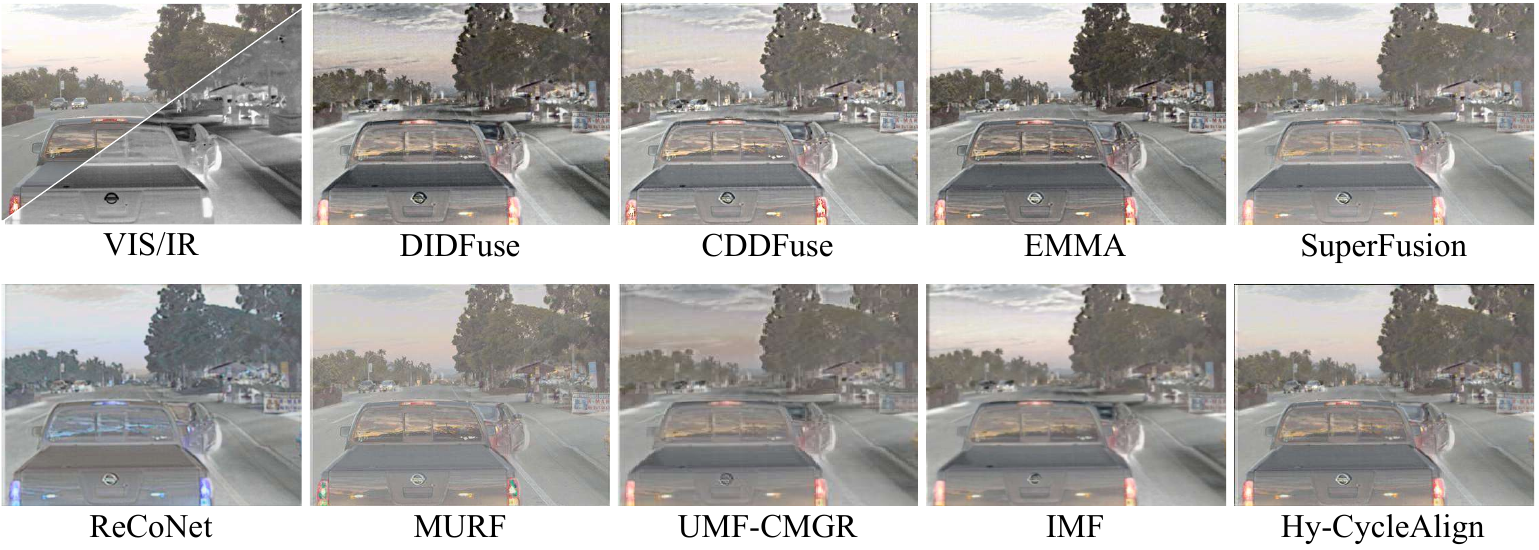}
   \caption{Comparison of results for the RoadScene dataset with rigid misalignment.}
   \label{result_rs_sup}
\end{figure}

\begin{table*}[th]
  \setlength\tabcolsep{3pt}
  \centering
   \caption{Quantitative comparisons at RoadScene and TNO. \textbf{Boldface} and \underline{underline} show the best and second-best values, respectively.}
   \label{tab:brats_isles} 
  
 \resizebox{\linewidth}{!}{%
\begin{tabular}{ c c c c c c c c c c c}
    \toprule
    \multirow{2}*{ \textbf{\textcolor{black}{Dataset}}}& \multirow{2}*{\textbf{\textcolor{black}{Metric}}}
    
    & \multicolumn{3}{c}{\textbf{\textcolor{black}{Alignment-free fusion methods}}}
    & \multicolumn{6}{c}{\textbf{\textcolor{black}{Alignment-based fusion methods}}}\\

     \cmidrule(r){3-5} \cmidrule(r){6-11}   
   {}&{}& \textbf{ \textcolor{black}{DIDFuse} } & \textbf{\textcolor{black}{CDDFuse}} & \textbf{\textcolor{black}{EMMA}} & \textbf{ \textcolor{black}{SuperFusion} } & \textbf{\textcolor{black}{ReCoNet}} & \textbf{\textcolor{black}{MURF}} & \textbf{\textcolor{black}{UMF-CMGR}} & \textbf{\textcolor{black}{IMF}} & \textbf{\textcolor{black}{Hy-CycleAlign}}\\
    \cmidrule(r){1-11}

    \multirow{7}*{\textbf{\textcolor{black}{\rotatebox{90}{RoadScene}}}} 
    & \textbf{HD}$\downarrow$ & $109.88$ & $\underline{99.34}$ &$\mathbf{91.63}$ &$\underline{81.56}$ & $87.42$ & $86.03$ &$105.02$ & $90.88$ & 	$\mathbf{80.24}$\\
    {}&\textbf{HD95}$\downarrow$ &	$69.04$ &$\underline{64.65}$ & $\mathbf{56.70}$ &$\underline{44.43}$ &$45.72$ & $39.36$ & $65.27$& $56.76$&	$\mathbf{34.94}$\\
    {}&\textbf{ASSD}$\downarrow$ &	$22.21$	& $\underline{18.10}$ & $\mathbf{15.26}$ &$10.38$&$11.57$& $\underline{8.43}$ &	$22.19$&	$18.32$&	$\mathbf{7.73}$\\
    {}&\textbf{DSC}$\uparrow$ &	$0.50$ &	$\underline{0.55}$ &$\mathbf{0.64}$  &$0.73$ & $\underline{0.74}$ & $0.60$ &$0.59$ & $0.60$&	$\mathbf{0.84}$\\
    {}&\textbf{MEE}$\downarrow$ & $27.24$ &	$\mathbf{14.76}$ & $\underline{25.42}$ &$\mathbf{10.06}$ & $17.01$ & $\underline{10.24}$ &	 $28.35$& $14.52$ &	$15.25$\\
    {}&\textbf{SF}$\uparrow$ &	$\underline{16.57}$ &$\mathbf{17.37}$ & $15.07$ &$\underline{11.67}$ &$9.02$ & $9.79$ & $3.86$&	$7.43$&	$\mathbf{12.3}$\\
    {}&\textbf{EN}$\uparrow$ & $\mathbf{7.50}$ &$7.27$ & $\underline{7.42}$ &$\mathbf{7.09}$ & $7.05$ & $6.73$ & $6.40$ &	 $7.04$ &	$\underline{7.06}$\\
    \cmidrule(r){1-11}

    \multirow{7}*{\textbf{\textcolor{black}{\rotatebox{90}{TNO}}}} 
    & \textbf{HD}$\downarrow$ & 	$\underline{102.22}$& 	$105.67$& 	$\mathbf{91.42}$&	$93.10$& 	$\mathbf{87.47}$& $111.18$ &	$91.00$&	$103.69$ & 	$\underline{87.87}$\\
    {}&\textbf{HD95}$\downarrow$&	$57.72$&	$\underline{56.12}$&	$\mathbf{51.55}$&	$47.47$&	$\underline{42.34}$ & $67.42$ &	$58.61$&	$64.16$&	$\mathbf{33.99}$\\
    {}&\textbf{ASSD}$\downarrow$&	$21.92$	&$\underline{18.48}$	&$\mathbf{15.39}$&	$13.57$&	$\underline{10.27}$& $19.42$ &	$19.21$&	$21.81$&	$\mathbf{8.56}$\\
    {}&\textbf{DSC}$\uparrow$&	$0.69$&	$\underline{0.73}$&	$\mathbf{0.82}$&	$0.79$ &	$\underline{0.83}$ & $0.73$ &	$0.58$ &	$0.59$&	$\mathbf{0.90}$\\
    {}&\textbf{MEE}$\downarrow$&	$29.94$&	$\mathbf{8.4}$&	$\underline{11.18}$&	$\underline{9.30}$&	$14.36$& $23.22$ &	$16.24$&	$11.90$&	$\mathbf{6.04}$\\
    {}&\textbf{SF}$\uparrow$&	$\underline{12.65}$&	$\mathbf{13.90}$&	$11.74$&	$\underline{9.47}$&	$7.96$& $3.53$ &	$7.05$&	$7.12$&	$\mathbf{9.72}$\\
    {}&\textbf{EN}$\uparrow$&	$6.85$&	$\underline{7.09}$&	$\mathbf{7.16}$&	$6.81$&	$6.68$&	$6.34$& $6.85$ &	$\underline{6.88}$&	$\mathbf{6.94}$\\

\bottomrule
\end{tabular}
}
\end{table*}

\subsection{More Downstream Results for Infrared-visible Applications}
We apply the registration and fusion results to the task of object detection from the viewpoint of drones. In this task, we use YOLO v11-m \cite{yolo11_ultralytics} as the detector and employ mAP@0.5, precision and recall as evaluation metrics. Compared with existing methods, Hy-CycleAlign achieves the highest detection precision and mAP, indicating that registration and fusion can effectively enhance the performance of object detection tasks. The results are shown in Fig. \ref{result_det2} and Tab. \ref{table_det2}.



\subsection{More Comparisons of Nonlinear Misalignment Fusion Results}
\begin{figure}[th]
  \centering
  \includegraphics[width=1\linewidth]{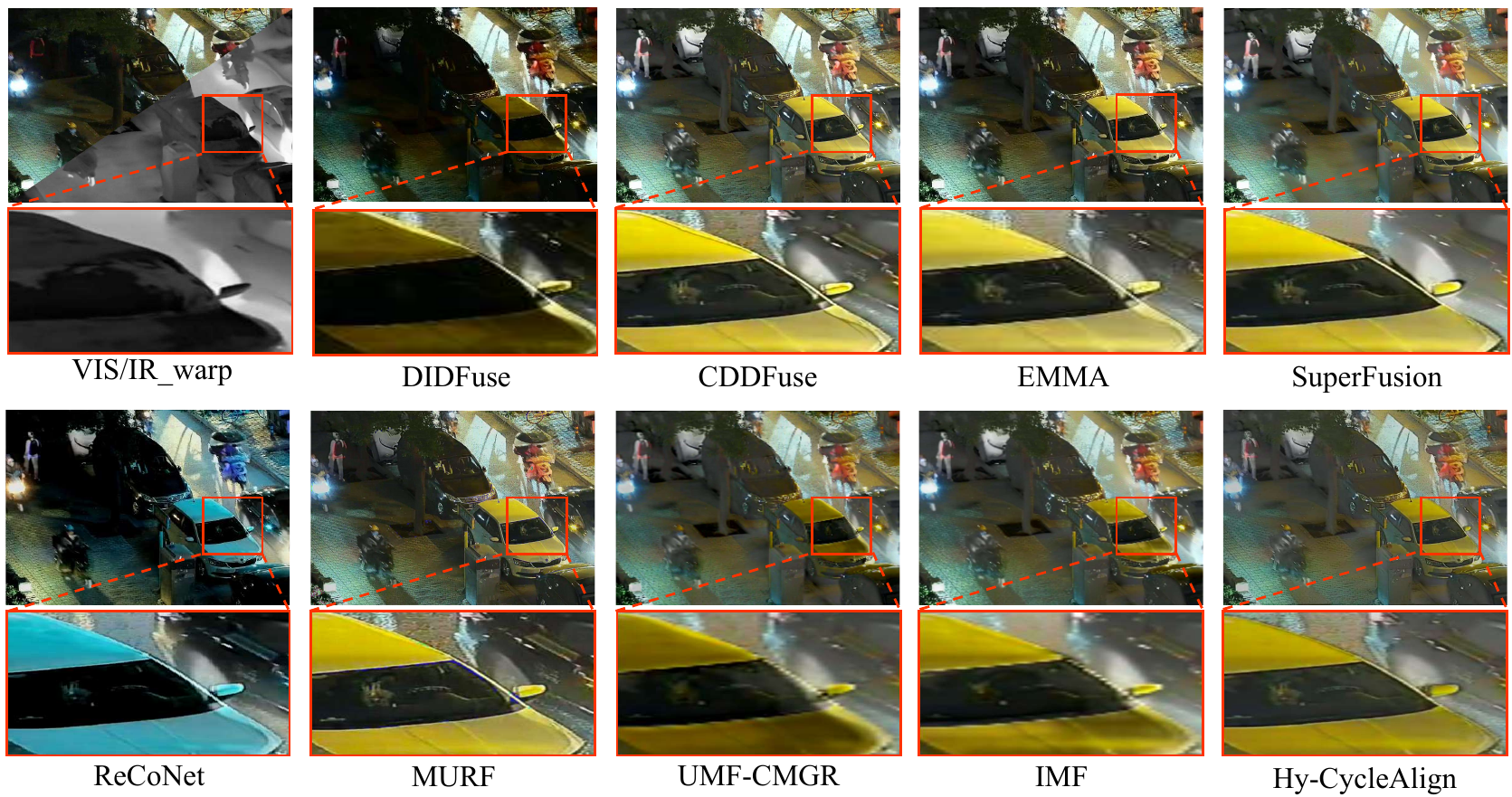}
   \caption{Comparison of results for the LLVIP dataset with infrared image nonlinear transformations.}
   \label{result_llvip_sup}
\end{figure}

\begin{figure}[th]
  \centering
  \includegraphics[width=1\linewidth]{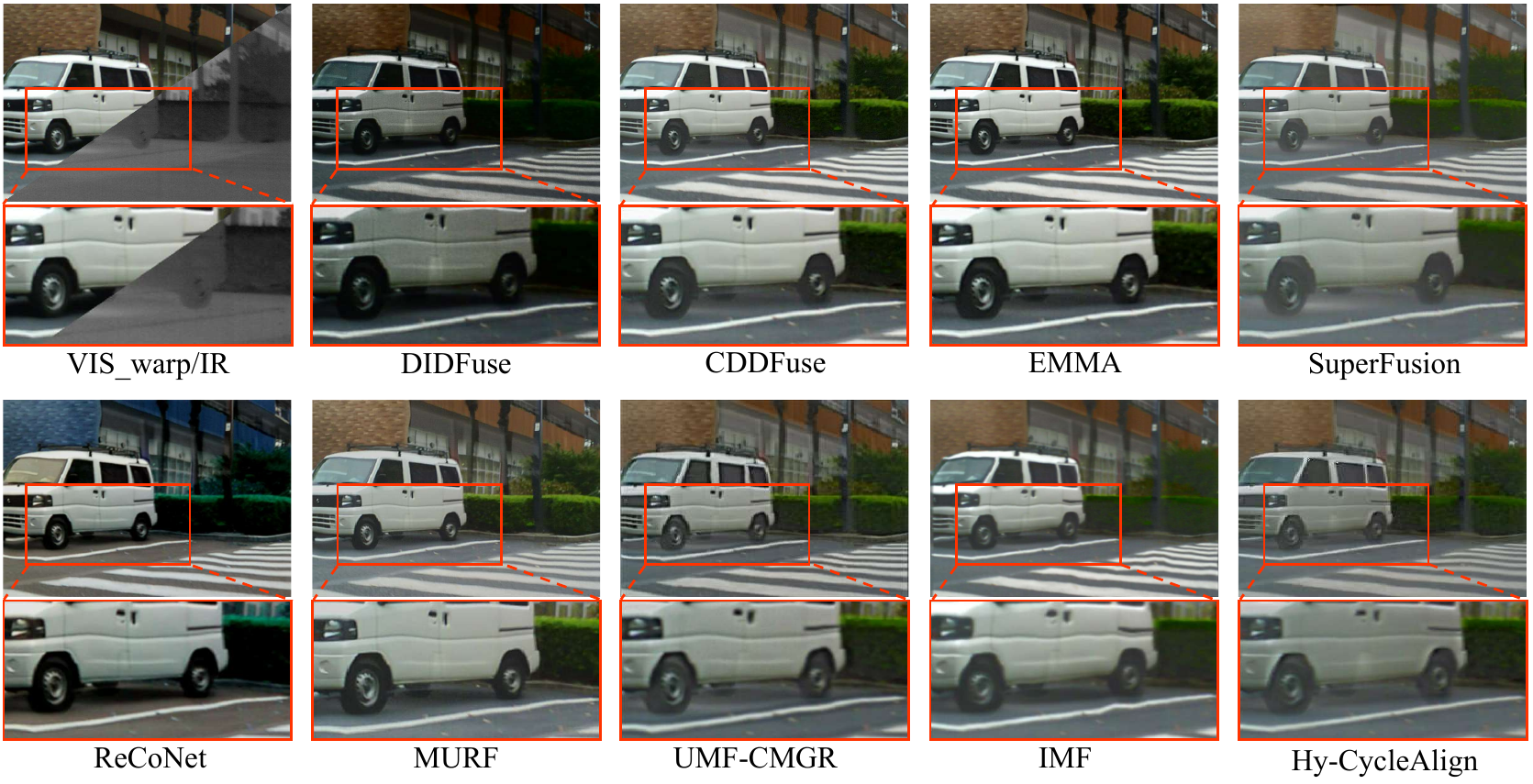}
   \caption{Comparison of results for the MFNet dataset with visible nonlinear transformations.}
   \label{result_mfnet_sup}
\end{figure}

Fig. \ref{result_llvip_sup} and \ref{result_mfnet_sup} present additional qualitative comparisons of infrared-visible image registration and fusion results. Our method handles misalignment more effectively while integrating thermal radiation from infrared images with texture details from visible images. Compared to other approaches, Hy-CycleAlign achieves more accurate multi-modal alignment under varying lighting conditions, better preserves fine textures, and highlights structural information.

\bibliographystyle{neurips_2025}  
\small
\bibliography{nips25}

\begin{thebibliography}{10}
\providecommand{\url}[1]{#1}
\csname url@samestyle\endcsname
\providecommand{\newblock}{\relax}
\providecommand{\bibinfo}[2]{#2}
\providecommand{\BIBentrySTDinterwordspacing}{\spaceskip=0pt\relax}
\providecommand{\BIBentryALTinterwordstretchfactor}{4}
\providecommand{\BIBentryALTinterwordspacing}{\spaceskip=\fontdimen2\font plus
\BIBentryALTinterwordstretchfactor\fontdimen3\font minus \fontdimen4\font\relax}
\providecommand{\BIBforeignlanguage}[2]{{%
\expandafter\ifx\csname l@#1\endcsname\relax
\typeout{** WARNING: IEEEtran.bst: No hyphenation pattern has been}%
\typeout{** loaded for the language `#1'. Using the pattern for}%
\typeout{** the default language instead.}%
\else
\language=\csname l@#1\endcsname
\fi
#2}}
\providecommand{\BIBdecl}{\relax}
\BIBdecl

\bibitem{zhang2020ifcnn}
Y.~Zhang, Y.~Liu, P.~Sun, H.~Yan, X.~Zhao, and L.~Zhang, ``Ifcnn: A general image fusion framework based on convolutional neural network,'' \emph{Information Fusion}, vol.~54, pp. 99--118, 2020.

\bibitem{xu2020u2fusion}
H.~Xu, J.~Ma, J.~Jiang, X.~Guo, and H.~Ling, ``U2fusion: A unified unsupervised image fusion network,'' \emph{IEEE Transactions on Pattern Analysis and Machine Intelligence}, vol.~44, no.~1, pp. 502--518, 2020.

\bibitem{zhao2020didfuse}
Z.~Zhao, S.~Xu, C.~Zhang, J.~Liu, P.~Li, and J.~Zhang, ``Didfuse: Deep image decomposition for infrared and visible image fusion,'' \emph{arXiv preprint arXiv:2003.09210}, 2020.

\bibitem{huang2022reconet}
Z.~Huang, J.~Liu, X.~Fan, R.~Liu, W.~Zhong, and Z.~Luo, ``Reconet: Recurrent correction network for fast and efficient multi-modality image fusion,'' in \emph{European conference on computer Vision}.\hskip 1em plus 0.5em minus 0.4em\relax Springer, 2022, pp. 539--555.

\bibitem{TANG2022SuperFusion_}
L.~Tang, Y.~Deng, Y.~Ma, J.~Huang, and J.~Ma, ``Superfusion: A versatile image registration and fusion network with semantic awareness,'' \emph{IEEE/CAA Journal of Automatica Sinica}, vol.~9, no.~12, pp. 2121--2137, 2022.

\bibitem{xu2023murf}
H.~Xu, J.~Yuan, and J.~Ma, ``Murf: Mutually reinforcing multi-modal image registration and fusion,'' \emph{IEEE transactions on pattern analysis and machine intelligence}, vol.~45, no.~10, pp. 12\,148--12\,166, 2023.

\bibitem{chen2022unsupervised}
Z.~Chen, J.~Wei, and R.~Li, ``Unsupervised multi-modal medical image registration via discriminator-free image-to-image translation,'' \emph{arXiv preprint arXiv:2204.13656}, 2022.

\bibitem{wang2024improving}
D.~Wang, J.~Liu, L.~Ma, R.~Liu, and X.~Fan, ``Improving misaligned multi-modality image fusion with one-stage progressive dense registration,'' \emph{IEEE Transactions on Circuits and Systems for Video Technology}, 2024.

\bibitem{bachmann2020constant}
G.~Bachmann, G.~B{\'e}cigneul, and O.~Ganea, ``Constant curvature graph convolutional networks,'' in \emph{International conference on machine learning}.\hskip 1em plus 0.5em minus 0.4em\relax PMLR, 2020, pp. 486--496.

\bibitem{chami2019hyperbolic}
I.~Chami, Z.~Ying, C.~R{\'e}, and J.~Leskovec, ``Hyperbolic graph convolutional neural networks,'' \emph{Advances in neural information processing systems}, vol.~32, 2019.

\bibitem{dai2021hyperbolic}
J.~Dai, Y.~Wu, Z.~Gao, and Y.~Jia, ``A hyperbolic-to-hyperbolic graph convolutional network,'' in \emph{Proceedings of the IEEE/CVF conference on computer vision and pattern recognition}, 2021, pp. 154--163.

\bibitem{cao2025hyperbolic}
H.~Cao, Y.~Wang, J.~Li, P.~Zhu, and Q.~Hu, ``Hyperbolic-euclidean deep mutual learning,'' in \emph{Proceedings of the ACM on Web Conference 2025}, 2025, pp. 3073--3083.

\bibitem{liu2019hyperbolic}
Q.~Liu, M.~Nickel, and D.~Kiela, ``Hyperbolic graph neural networks,'' \emph{Advances in neural information processing systems}, vol.~32, 2019.

\bibitem{lou2020differentiating}
A.~Lou, I.~Katsman, Q.~Jiang, S.~Belongie, S.-N. Lim, and C.~De~Sa, ``Differentiating through the fr{\'e}chet mean,'' in \emph{International conference on machine learning}.\hskip 1em plus 0.5em minus 0.4em\relax PMLR, 2020, pp. 6393--6403.

\bibitem{aly2020every}
R.~Aly, A.~Ossa, A.~K{\"o}hn, C.~Biemann, A.~Panchenko, and S.~Acharya, ``Every child should have parents: A taxonomy refinement algorithm based on hyperbolic term embeddings,'' in \emph{ACL 2019-57th Annual Meeting of the Association for Computational Linguistics, Proceedings of the Conference}, 2020, pp. 4811--4817.

\bibitem{tifrea2018poincar}
A.~Tifrea, G.~B{\'e}cigneul, and O.-E. Ganea, ``Poincar{\'e} glove: Hyperbolic word embeddings,'' \emph{arXiv preprint arXiv:1810.06546}, 2018.

\bibitem{zhu2020hypertext}
Y.~Zhu, D.~Zhou, J.~Xiao, X.~Jiang, X.~Chen, and Q.~Liu, ``Hypertext: Endowing fasttext with hyperbolic geometry,'' \emph{arXiv preprint arXiv:2010.16143}, 2020.

\bibitem{ramasinghe2024accept}
S.~Ramasinghe, V.~Shevchenko, G.~Avraham, and A.~Thalaiyasingam, ``Accept the modality gap: An exploration in the hyperbolic space,'' in \emph{Proceedings of the IEEE/CVF Conference on Computer Vision and Pattern Recognition}, 2024, pp. 27\,263--27\,272.

\bibitem{yang2024hypformer}
M.~Yang, H.~Verma, D.~C. Zhang, J.~Liu, I.~King, and R.~Ying, ``Hypformer: Exploring efficient transformer fully in hyperbolic space,'' in \emph{Proceedings of the 30th ACM SIGKDD Conference on Knowledge Discovery and Data Mining}, 2024, pp. 3770--3781.

\bibitem{mettes2024hyperbolic}
P.~Mettes, M.~Ghadimi~Atigh, M.~Keller-Ressel, J.~Gu, and S.~Yeung, ``Hyperbolic deep learning in computer vision: A survey,'' \emph{International Journal of Computer Vision}, vol. 132, no.~9, pp. 3484--3508, 2024.

\bibitem{li2024hyperbolic}
H.~Li, Z.~Chen, Y.~Xu, and J.~Hu, ``Hyperbolic anomaly detection,'' in \emph{Proceedings of the IEEE/CVF Conference on Computer Vision and Pattern Recognition}, 2024, pp. 17\,511--17\,520.

\bibitem{khrulkov2020hyperbolic}
V.~Khrulkov, L.~Mirvakhabova, E.~Ustinova, I.~Oseledets, and V.~Lempitsky, ``Hyperbolic image embeddings,'' in \emph{Proceedings of the IEEE/CVF conference on computer vision and pattern recognition}, 2020, pp. 6418--6428.

\bibitem{atigh2022hyperbolic}
M.~G. Atigh, J.~Schoep, E.~Acar, N.~Van~Noord, and P.~Mettes, ``Hyperbolic image segmentation,'' in \emph{Proceedings of the IEEE/CVF conference on computer vision and pattern recognition}, 2022, pp. 4453--4462.

\bibitem{fu2024cf}
H.~Fu, J.~Yuan, G.~Zhong, X.~He, J.~Lin, and Z.~Li, ``Cf-deformable detr: an end-to-end alignment-free model for weakly aligned visible-infrared object detection,'' in \emph{Proceedings of the Thirty-Third International Joint Conference on Artificial Intelligence}, 2024, pp. 758--766.

\bibitem{brown1992survey}
L.~G. Brown, ``A survey of image registration techniques,'' \emph{ACM computing surveys (CSUR)}, vol.~24, no.~4, pp. 325--376, 1992.

\bibitem{kong2021breaking}
L.~Kong, C.~Lian, D.~Huang, Y.~Hu, Q.~Zhou \emph{et~al.}, ``Breaking the dilemma of medical image-to-image translation,'' \emph{Advances in Neural Information Processing Systems}, vol.~34, pp. 1964--1978, 2021.

\bibitem{wu2021discover}
Q.~Wu, P.~Dai, J.~Chen, C.-W. Lin, Y.~Wu, F.~Huang, B.~Zhong, and R.~Ji, ``Discover cross-modality nuances for visible-infrared person re-identification,'' in \emph{Proceedings of the IEEE/CVF conference on computer vision and pattern recognition}, 2021, pp. 4330--4339.

\bibitem{fan2024gls}
Z.~Fan, Y.~Pi, M.~Wang, Y.~Kang, and K.~Tan, ``Gls--mift: A modality invariant feature transform with global-to-local searching,'' \emph{Information Fusion}, vol. 105, p. 102252, 2024.

\bibitem{xiang2022infrared}
L.~Xiang, L.~Zhao, S.~Chen, and X.~Li, ``Infrared and visible image registration in uav inspection,'' in \emph{Proceedings of the 2022 6th International Conference on Video and Image Processing}, 2022, pp. 67--71.

\bibitem{itti2002model}
L.~Itti, C.~Koch, and E.~Niebur, ``A model of saliency-based visual attention for rapid scene analysis,'' \emph{IEEE Transactions on pattern analysis and machine intelligence}, vol.~20, no.~11, pp. 1254--1259, 2002.

\bibitem{li2023efficient}
Y.~Li, Y.~Fan, X.~Xiang, D.~Demandolx, R.~Ranjan, R.~Timofte, and L.~Van~Gool, ``Efficient and explicit modelling of image hierarchies for image restoration,'' in \emph{Proceedings of the IEEE/CVF Conference on Computer Vision and Pattern Recognition}, 2023, pp. 18\,278--18\,289.

\bibitem{pal2024compositional}
A.~Pal, M.~van Spengler, G.~M.~D. di~Melendugno, A.~Flaborea, F.~Galasso, and P.~Mettes, ``Compositional entailment learning for hyperbolic vision-language models,'' \emph{arXiv preprint arXiv:2410.06912}, 2024.

\bibitem{zhu2017unpaired}
J.-Y. Zhu, T.~Park, P.~Isola, and A.~A. Efros, ``Unpaired image-to-image translation using cycle-consistent adversarial networks,'' in \emph{Proceedings of the IEEE international conference on computer vision}, 2017, pp. 2223--2232.

\bibitem{tang2022image}
L.~Tang, J.~Yuan, and J.~Ma, ``Image fusion in the loop of high-level vision tasks: A semantic-aware real-time infrared and visible image fusion network,'' \emph{Information Fusion}, vol.~82, pp. 28--42, 2022.

\bibitem{zhao2023cddfuse}
Z.~Zhao, H.~Bai, J.~Zhang, Y.~Zhang, S.~Xu, Z.~Lin, R.~Timofte, and L.~Van~Gool, ``Cddfuse: Correlation-driven dual-branch feature decomposition for multi-modality image fusion,'' in \emph{Proceedings of the IEEE/CVF conference on computer vision and pattern recognition}, 2023, pp. 5906--5916.

\bibitem{sun2022drone}
Y.~Sun, B.~Cao, P.~Zhu, and Q.~Hu, ``Drone-based rgb-infrared cross-modality vehicle detection via uncertainty-aware learning,'' \emph{IEEE TCSVT}, vol.~32, no.~10, pp. 6700--6713, 2022.

\bibitem{jia2021llvip}
X.~Jia, C.~Zhu, M.~Li, W.~Tang, and W.~Zhou, ``Llvip: A visible-infrared paired dataset for low-light vision,'' in \emph{Proceedings of the IEEE/CVF international conference on computer vision}, 2021, pp. 3496--3504.

\bibitem{ha2017mfnet}
Q.~Ha, K.~Watanabe, T.~Karasawa, Y.~Ushiku, and T.~Harada, ``Mfnet: Towards real-time semantic segmentation for autonomous vehicles with multi-spectral scenes,'' in \emph{2017 IEEE/RSJ International Conference on Intelligent Robots and Systems (IROS)}.\hskip 1em plus 0.5em minus 0.4em\relax IEEE, 2017, pp. 5108--5115.

\bibitem{loshchilov2017fixing}
I.~Loshchilov, F.~Hutter \emph{et~al.}, ``Fixing weight decay regularization in adam,'' \emph{arXiv preprint arXiv:1711.05101}, vol.~5, 2017.

\bibitem{Zhao_2024_CVPR}
Z.~Zhao, H.~Bai, J.~Zhang, Y.~Zhang, K.~Zhang, S.~Xu, D.~Chen, R.~Timofte, and L.~Van~Gool, ``Equivariant multi-modality image fusion,'' in \emph{Proceedings of the IEEE/CVF Conference on Computer Vision and Pattern Recognition (CVPR)}, June 2024, pp. 25\,912--25\,921.

\bibitem{wang2022unsupervised}
D.~Wang, J.~Liu, X.~Fan, and R.~Liu, ``Unsupervised misaligned infrared and visible image fusion via cross-modality image generation and registration,'' \emph{arXiv preprint arXiv:2205.11876}, 2022.

\bibitem{jiang2023breaking}
Z.~Jiang, Z.~Zhang, J.~Liu, X.~Fan, and R.~Liu, ``Breaking modality disparity: Harmonized representation for infrared and visible image registration,'' \emph{arXiv preprint arXiv:2304.05646}, 2023.

\bibitem{ganea2018hyperbolic}
O.~Ganea, G.~B{\'e}cigneul, and T.~Hofmann, ``Hyperbolic neural networks,'' \emph{Advances in neural information processing systems}, vol.~31, 2018.

\bibitem{zhang2023multimodal}
L.~Zhang, S.~Na, T.~Liu, D.~Zhu, and J.~Huang, ``Multimodal deep fusion in hyperbolic space for mild cognitive impairment study,'' in \emph{International Conference on Medical Image Computing and Computer-Assisted Intervention}.\hskip 1em plus 0.5em minus 0.4em\relax Springer, 2023, pp. 674--684.

\bibitem{simonyan2014very}
K.~Simonyan and A.~Zisserman, ``Very deep convolutional networks for large-scale image recognition,'' \emph{arXiv preprint arXiv:1409.1556}, 2014.

\bibitem{yolo11_ultralytics}
\BIBentryALTinterwordspacing
G.~Jocher and J.~Qiu, ``Ultralytics yolo11,'' 2024. [Online]. Available: \url{https://github.com/ultralytics/ultralytics}
\BIBentrySTDinterwordspacing

\bibitem{xu2020aaai}
H.~Xu, J.~Ma, Z.~Le, J.~Jiang, and X.~Guo, ``Fusiondn: A unified densely connected network for image fusion,'' in \emph{proceedings of the Thirty-Fourth AAAI Conference on Artificial Intelligence}, 2020.

\bibitem{toet2012progress}
A.~Toet and M.~A. Hogervorst, ``Progress in color night vision,'' \emph{Optical Engineering}, vol.~51, no.~1, pp. 010\,901--010\,901, 2012.

\end{thebibliography}
\normalsize

\end{document}